\documentclass[10pt,twocolumn,letterpaper]{article}

\usepackage[pagenumbers]{cvpr}

\graphicspath{{Figures/}{images/}{imgs/}{resources}}
\usepackage{algorithm}
\usepackage{algpseudocode}
\floatname{algorithm}{Algorithm}

\usepackage{microtype}

\usepackage{dsfont}
\usepackage{nicefrac}

\usepackage{multirow}
\usepackage{colortbl}
\usepackage{overpic}
\usepackage[frozencache,cachedir=.]{minted}

\usepackage{xcolor}

\definecolor{eins}{RGB}{102,194,164}
\definecolor{zwei}{RGB}{178,226,226}
\definecolor{drei}{RGB}{237,248,251}
\definecolor{einsText}{RGB}{57,146,116}

\newcommand{\tabbox}[1]{\makebox[1em][l]{#1}}
\newcommand{\myparagraph}[1]{\vspace{2pt}\noindent\textbf{#1}\hspace{1pt}}

\definecolor{cvprblue}{rgb}{0.21,0.49,0.74}
\usepackage[pagebackref,breaklinks,colorlinks,citecolor=cvprblue]{hyperref}

\title{INPC: Implicit Neural Point Clouds for Radiance Field Rendering
\vspace{-15pt}
}

\author{
Florian Hahlbohm$^1$\qquad
Linus Franke$^2$\qquad
Moritz Kappel$^1$\\
Susana Castillo$^1$\qquad
Martin Eisemann$^1$\qquad
Marc Stamminger$^2$\qquad
Marcus Magnor$^1$\\
{\parbox{\textwidth}{
\centering \small $^1$ Computer Graphics Lab, TU Braunschweig, Germany
\hspace{7pt}{\tt\small \{lastname\}@cg.tu-bs.de}\\
\small $^2$ Visual Computing Erlangen, FAU Erlangen-Nürnberg, Germany
\hspace{7pt}{\tt\small \{firstname.lastname\}@fau.de}}}
}

\begin{document}

\twocolumn[{ 
\renewcommand\twocolumn[1][]{#1} 
\maketitle 
\centering
\vspace{-0.7cm} 
\includegraphics[width=0.8\textwidth, keepaspectratio]{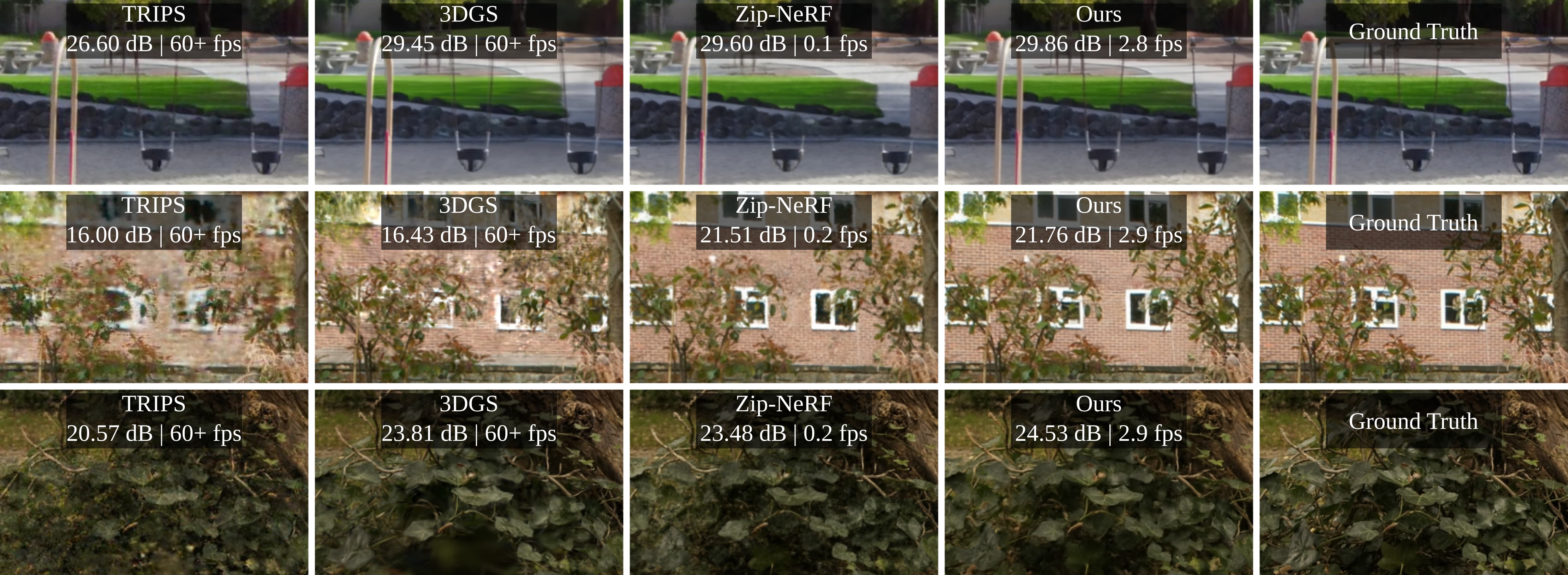}
\captionof{figure}{Novel views synthesized by our model and three state-of-the-art baselines. Our implicit point cloud optimization excels at capturing fine detail leading to a higher visual fidelity compared to baselines. While outperformed by explicit point-based methods~\cite{kerbl3Dgaussians, franke2024trips} in terms of inference frame rates, our model renders 17$\times$ faster than Zip-NeRF~\cite{barron2023ICCV}. Per-patch PSNR and per-scene fps values are inset.\vspace{0.3cm} 
}\label{fig:teaser} 
}]

\begin{abstract}
We introduce a new approach for reconstruction and novel view synthesis of unbounded real-world scenes.
In contrast to previous methods using either volumetric fields, grid-based models, or discrete point cloud proxies, we propose a hybrid scene representation, which \textit{implicitly} encodes the geometry in a continuous octree-based probability field and view-dependent appearance in a multi-resolution hash grid.
This allows for extraction of arbitrary \textit{explicit} point clouds, which can be rendered using rasterization.
In doing so, we combine the benefits of both worlds and retain favorable behavior during optimization:
Our novel implicit point cloud representation and differentiable bilinear rasterizer enable fast rendering while preserving the fine geometric detail captured by volumetric neural fields.
Furthermore, this representation does not depend on priors like structure-from-motion point clouds.
Our method achieves state-of-the-art image quality on common benchmarks.
Furthermore, we achieve fast inference at interactive frame rates, and can convert our trained model into a large, explicit point cloud to further enhance performance.
\end{abstract}

\section{Introduction}
\label{sec:intro}
Novel view synthesis describes the task of rendering novel inter- or extrapolated views from a set of input images, which is an inherently difficult problem.
Recent methods commonly address this by reconstructing the scene either volumetrically as dense implicit radiance fields~\cite{penner2017soft} or use explicit geometric representations~\cite{shum2000review} such as point clouds or meshes.
Leveraging advances in optimization-based neural rendering techniques, volumetric approaches achieve an impressive increase in quality by optimizing geometric information (i.e., density and appearance) into \textit{Multi-Layer Perceptrons (MLPs)}~\cite{mildenhall2020nerf}, voxel grids~\cite{yu2021plenoxels}, or hash maps~\cite{mueller2022instant}.
In contrast, explicit point-based methods optimize appearance at discrete 3D scene points, where color is being represented via 3D Gaussians~\cite{kerbl3Dgaussians} or per-point features~\cite{aliev2020npbg}.
While both directions provide unique advantages, they also entail severe drawbacks such as volumetric methods relying on computationally intensive ray-marching or explicit methods requiring a-priori point proxies.

In this work, we aim at creating a novel, efficient, and robust scene representation which combines the benefits of both worlds while bypassing the need for expensive ray-casting or explicit priors.
To this end, we introduce Implicit Neural Point Cloud (INPC), an implicit scene representation that enables accurate scene reconstruction by sampling and subsequent rendering of explicit point clouds using fast differentiable rasterization.
More specifically, we take inspiration from current state-of-the-art approaches 3D Gaussian Splatting~\cite{kerbl3Dgaussians} and Zip-NeRF~\cite{barron2023ICCV}:
We decompose and optimize a scene into two parts, which constitutes a concept that we dub \textit{implicit point clouds}.
Here, the geometric structure is represented as an octree-based point probability field, while appearance features are embedded in an \textit{implicit} coordinate-based multi-resolution hash grid.
The octree is progressively subdivided to ensure similar probability distribution across all leaf nodes, which enables the reconstruction of fine articulated geometry while maximizing the capacity of our appearance hash grid.
During rendering, we use the probability octree as an estimator for point positions and use either random positions in each leaf or fixed sampling patterns, while per-point appearance features are queried from the hash grid.
The resulting \textit{explicit} point cloud is then rendered via fast bilinear splatting, where gradients are backpropagated through our differentiable end-to-end pipeline to the implicit representation.

Notably, our formulation makes use of the favorable optimization properties of volumetric methods by implicitly modeling geometry and appearance.
Simultaneously, it elegantly replaces interval-based sampling of per-pixel rays with a unified sampling step for the whole frustum of a given viewpoint.
This subsequently allows for forward rendering with a rasterizer, i.e., the driving factor for real-time frame rates of point-based methods.
By combining the benefits of both families of approaches, our method achieves robust radiance field reconstruction alongside rendering with state-of-the-art quality on benchmark datasets.
In summary, our contributions are:
\begin{itemize}
    \item The introduction of implicit neural point clouds as a data structure to effectively reconstruct unbounded 3D scenes.
    \item An algorithm for extracting view-specific point clouds as well as global point clouds from this model.
    \item A fast and differentiable rendering formulation for this data structure using bilinearly splatted points.
\end{itemize}

\section{Related Work}
Traditionally, novel view synthesis was based on light fields~\cite{gortler1996lumigraph}, however image-based rendering became a popular alternative~\cite{shum2000review}.
It commonly works by warping source views onto geometric proxies~\cite{debevec1998efficient, chaurasia2013depth}.
This proxy may contain artifacts, especially near object edges, either due to limited input coverage or misaligned cameras.
With image-based rendering, these artifacts result in blurred and inaccurate images, with subsequent methods lessening these artifacts~\cite{eisemann2008floating,chaurasia2013depth}.
The geometric proxy can also be a full 3D reconstruction, which with the introduction of \textit{Structure-from-Motion (SfM)}~\cite{schoenberger2016colmap1} and \textit{Multi-View Stereo (MVS)}~\cite{seitz2006comparison} gained popularity.
Furthermore, the advent of deep learning-based techniques in this field further improved results~\cite{Tewari2022NeuRendSTAR} through learned blending operators~\cite{hedman2018deep} and textures~\cite{thies2019deferred}, lessening failure cases introduced by artifacts in the reconstruction.
In the following, we discuss related works in volume- and point-based novel view synthesis, the two directions we combine in our work.

\begin{figure*}[t]
\centering
\includegraphics[width=.9\linewidth, keepaspectratio]{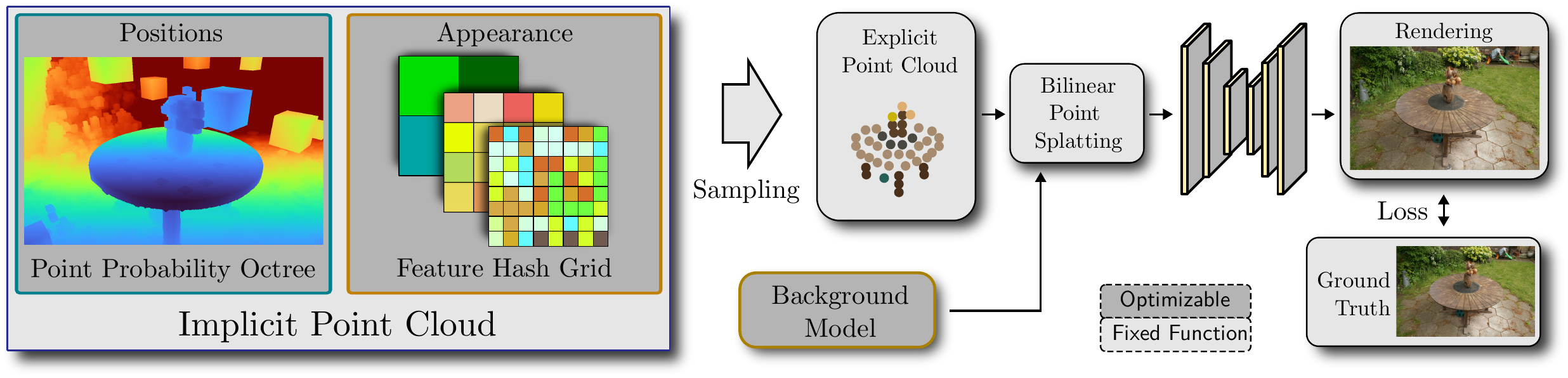}%
\caption{
\textbf{Overview of our method:} We introduce the implicit point cloud, a combination of a point probability field stored in an octree and implicitly stored appearance features. 
To render an image for a given viewpoint, we sample the representation by estimating point positions and querying the multi-resolution hash grid for per-point features. 
This explicit point cloud -- together with a small background MLP -- is then rendered with a bilinear point splatting module and post-processed by a CNN.
During optimization, the neural networks as well as the implicit point cloud are optimized, efficiently reconstructing the scene.
}\label{fig:pipeline}
\end{figure*}

\paragraph{Neural Radiance Fields.}
Recently, implicitly representing 3D scenes within volumetric fields became popular, enabling novel view synthesis via volume rendering and without the need for proxy geometry.
\citet{mildenhall2020nerf} showed exceptional results by compressing a complete 3D scene into a large coordinate-based MLP, a concept called \textit{Neural Radiance Field (NeRF)}.
To render images, pixel-wise ray-marching is used with the volume rendering formulation for each pixel color $C$:
\begin{equation}
    \label{eq:alpha_blend_points}
	C = \sum_{i=1}^N T_i\alpha_i\mathbf{c}_i \hspace{1.5em} \text{and} \hspace{1.5em}
	T_i = \prod_{j=1}^{i-1}(1-\alpha_j),
\end{equation}
with $\alpha_i = 1-e^{-\sigma_i\delta_i}$. 
Here, density $\sigma_i$ and color $\mathbf{c}_i$ are the outputs of the MLP at each ray interval $\delta_i$.
Several successive works aim to resolve challenges bound to this concept by addressing input view distributions~\cite{chibane2021stereo,yu2021pixelnerf,kopanas2023improving,wu2024reconfusion} and computation times~\cite{barron2021mipnerf,neff2021donerf,chen2021mvsnerf,chibane2021stereo, mueller2022instant,tancik2021learned,turki2022mega}.
For the latter, discretizing the scene space using voxel grids~\cite{yu2021plenoxels}, octrees~\cite{ruckert2022neat,yu2021plenoctrees}, tensor decomposition~\cite{chen2022tensorf,reiser2023merf,duckworth2023smerf}, or even distilling a faster model for inference with neurally textured triangle meshes~\cite{chen2023mobilenerf} or mesh baking~\cite{yariv2023bakedsdf,reiser2024binary} proved effective.
In terms of training and rendering speed, \textit{Instant-NGP}~\cite{mueller2022instant} presented exceptional results with a hash grid-based space partitioning scheme, allowing training within minutes and frame rates of up to 10 fps.
Furthermore, close to our approach are hybrids of conventional and optimized ideas: Point-NeRF~\cite{xu2022point} uses an explicit point cloud with neural features, however, images are rendered with slow ray-marching and are restricted to reconstruct bounded scenes. 
Regarding quality, \citet{barron2021mipnerf} propose anti-aliasing through conical frustum sampling and enable extension to unbounded scenes via space contraction~\cite{barron2022mipnerf360}.
The current state of the art in this field, \textit{Zip-NeRF}~\cite{barron2023ICCV}, combines ideas from both quality and efficiency directions.
It augments the underlying grid-based data structure with anti-aliasing through conical sampling, addition of scaling information, and refined empty space skipping.
While {Zip-NeRF} is comparably fast in training (taking about 5 hours), the rendering speed for novel views is limited to $\sim$0.2 fps on consumer-grade hardware.
This approach is close to our method, as we also recombine grid-based appearance information with implicit density formulation for scene reconstruction.
In contrast to {Zip-NeRF}, however, our method enables rasterization-based rendering, making it faster in inference.

\paragraph{Point Rendering.}
Orthogonal to NeRFs, neurally rendering radiance fields via explicit points is an established and efficient methodology for novel view synthesis.
Early work builds on established techniques~\cite{hornung2009interactive} and associates point clouds obtained through MVS with optimizable colors or features~\cite{aliev2020npbg}. 
Points are rendered as splats of varying sizes and a \textit{convolutional neural network (CNN)} is used to interpret features.
Due to the unstructured and disconnected nature of point clouds, holes in image space where no point was projected to can appear, which the neural network can also resolve.
Several follow-up works have been proposed eliminating training time~\cite{rakhimov2022npbgpp,harrerfranke2023inovis}, adding per-view feature optimization~\cite{kopanas2021perviewopt}, reflection warp-fields~\cite{kopanas2022catacaustics}, or differentiable tone mapping~\cite{ruckert2022adop}.
Furthermore, differentiability with respect to point positions and camera parameters -- via approximate~\cite{ruckert2022adop} or linear gradients~\cite{franke2024trips} -- has been introduced, retaining rendering performance for large point clouds compared to other point representations~\cite{lassner2021pulsar,yifan2019differentiable,wiles2020synsin}.
Recent point-based radiance field renderers~\cite{kopanas2021perviewopt,kopanas2022catacaustics,franke2023vet,franke2024trips,yifan2019differentiable,kerbl3Dgaussians} also adapt NeRFs rendering technique to $\alpha$-blending of points. 
Instead of taking $N$ samples along a ray as in \cref{eq:alpha_blend_points}, they blend together $\mathcal{N}$ sorted points with associated colors $c$ and opacities $\alpha$ instead of computing $\alpha$ based on density.
For approaches keeping true to MVS reconstructions, \textit{Trilinear Point Splatting (TRIPS)}~\cite{franke2024trips} is closest to our point rendering, as it also uses bilinear point splatting followed by a neural network for feature decoding.
However its reconstruction quality is limited by the quality of the MVS reconstruction, as no point cloud augmentations are performed.

Overcoming the reliance on point cloud priors has been proposed in several ways.
Recent improvements to MVS algorithms via CNNs~\cite{yao2018mvsnet,vats2024gc} as well as leveraging NeRFs~\cite{wei2021nerfingmvs} or transformers~\cite{ding2022transmvsnet,wang2022mvster} lessened the problem, while recent approaches also included the point cloud optimization process directly into the radiance field rendering pipeline.
This is done either with additional points via error propagation~\cite{zuo2023snp}, 3D error volumes~\cite{franke2023vet}, point growing via density estimation~\cite{xu2022point}, or gradient-based densification~\cite{kerbl3Dgaussians}.
Especially important here is \textit{3D Gaussian Splatting (3DGS)}~\cite{kerbl3Dgaussians}, which extends point rendering with anisotropic 3D Gaussians as a radiance field rendering paradigm.
Since its publication, it has become the basis for a multitude of follow-up works~\cite{yu2024mip, guedon2024sugar, huang20242d, radl2024stopthepop, kerbl2024hierarchical, wu2024recent}.
Apart from removing the need for a CNN to fill holes, the \textit{3DGS} formulation allows starting with a sparser point cloud that is densified by repeatedly splitting large Gaussians during the optimization.
In contrast, our approach captures more fine details, as we are able to optimize and render more detailed geometric and appearance information in our implicit point cloud. Furthermore, we are independent from an initial point cloud, thus increasing robustness. 

\section{Method}
\label{sec:method}
A minimal point cloud capable of novel view synthesis, is described by multiple data points consisting of a 3D position vector as well as color information, either though explicit RGB colors or decodable neural features.
In our method, instead of storing these spatial and photometric properties in the same data structure, we split them and optimize both \textit{implicitly}: Positions as an octree-based probability field $\mathcal{P}$ and appearance, i.e., colors/features and opacity, as a neural field $\mathcal{A}$. 
These two parts combined constitute what we call an \textit{implicit neural point cloud}.

Input to our method is a collection of RGB images with poses and an estimated bounding box.
From this, we initialize $\mathcal{P}$ as a voxel-based structure, which our algorithm iteratively refines into an octree to store probabilities for geometry (\cref{ssec:sampling_octree}).
This structure is then used as an estimator for point positions (\cref{ssec:point_sampling_strategies}).
These positions are fed into $\mathcal{A}$, for which we use a multi-resolution hash grid~\cite{mueller2022instant}, to retrieve opacity and spherical harmonics (SH) appearance features (\cref{ssec:hash_grid_ar}).
These parts are core to our proposed implicit point cloud structure, allowing us to optimize a radiance field efficiently, with fine geometric granularity, and great detail.

An overview of our method is depicted in \cref{fig:pipeline}.
For a given viewpoint, we first obtain a set of point positions using $\mathcal{P}$ and query $\mathcal{A}$ to retrieve per-point appearance features.
The resulting explicit point cloud is then rendered by first rasterizing with bilinear point splatting (\cref{ssec:differentiable_bilinear_point_splatting}) into a 2D feature image, then $\alpha$-composited with the output of a background model, and post-processed by a rendering network with a U-Net architecture~\cite{ronneberger2015unet} (\cref{ssec:neural_pp_network}).
During training, our model is optimized end-to-end with losses and regularizers carefully selected for our method (\cref{ssec:optimization_loss}).
We proceed by describing all components in detail.

\subsection{Sparse Point Probability Octree}
\label{ssec:sampling_octree}
Existing NeRF methods do not require explicit geometry as they place point samples along per-pixel camera rays~\cite{mildenhall2020nerf}.
While prior work demonstrated that this sampling scheme leads to aliasing artifacts~\cite{barron2021mipnerf}, NeRF-based methods can leverage the favorable optimization properties of implicit volumetric models to achieve state-of-the-art image quality~\cite{barron2023ICCV}.
In contrast, point-based methods require a persistent set of explicit point positions during the optimization.
Through optimization of point positions with approximated gradients~\cite{ruckert2022adop} as well as handcrafted splitting, merging, and pruning operations~\cite{kerbl3Dgaussians, franke2023vet}, these methods aim to refine an initial point cloud that is provided, e.g., as a byproduct of SfM algorithms.
Importantly, the approximated position gradients as well as heuristics used for refining the point cloud, lead to a less robust optimization compared to that of implicit models.

One goal of this work is removing the need for a persistent set of point positions within a typical point-based neural rendering pipeline.
In prior work, regular occupancy grids have proven to be a useful tool for skipping empty space~\cite{mueller2022instant,yu2021plenoxels}.
Notably, the tracked occupancy is closely related to the volume density used to represent scene geometry in volumetric NeRF-based models.
Our key observation is that this value can be normalized across the whole scene and interpret as a probability for geometry.
However, as the fixed voxel size of regular occupancy grids makes it difficult to accurately model complex geometry, we propose using a sparse octree to store a point probability $p_i$ for each leaf node.
During the optimization, nodes are updated, subdivided, or pruned, which allows for accurate reconstruction of fine geometric structures.
By sampling the multinomial distribution represented by this octree, we can extract point clouds on demand (\cref{ssec:point_sampling_strategies}).

\paragraph{Initialization.}
We initialize the probability field with a uniform 3D voxel grid as the initial leaf nodes of the sparse octree with probabilities set to a uniform value to achieve equal sampling probability.
\textit{Optionally}, we incorporate a point cloud as a prior, which slightly improves initial convergence of the optimization (see \cref{ssec:appendix_octree_details} for details).

\paragraph{Probability Updates.}
\label{par:updates}
Theoretically, the probability of each leaf node should represent how much of its volume is occupied by geometry \textit{relative to other leaf nodes}.
To this end, we employ an updating strategy inspired by occupancy grid updates in Instant-NGP~\cite{mueller2022instant} that combines exponential decay and knowledge from the current 3D model.
During optimization, we update the probabilities of all leaf nodes ($p_i$) after each optimization step using the following formula:
\begin{equation}
\label{eq:weight_update}
    p_i = \max{(\lambda_u \cdot p_{i}, \max{(\{b_0,...,b_n\})})},
\end{equation}
where $\{b_0,...,b_n\}$ is the set of $\alpha$-blending weights of all $n$ points extracted from the $i$-th leaf node.
Note that we use the transmitted blending weight $b_j=\alpha_jT_j$ (\cref{eq:alpha_blend_points}) computed by our rasterizer for the updates.
This imposes a visibility prior on the probabilities, as occluded points receive less transmittance.
For the exponential decay, we found $\lambda_u=0.9968$ to work well in practice and use it in all experiments.
We also skip updates during the first $100$ iterations for increased robustness, as $\mathcal{A}$ does not contain reliable information yet.

\paragraph{Subdivision.}
To refine the represented geometry, we want to iteratively subdivide and prune empty space in leaf nodes.
Intuitively, we achieve this by identifying partially unoccupied voxels, which we locate by tracking the difference between the largest and smallest blending weight of all $n$ points extracted from the $i$-th leaf node in each iteration:
\begin{equation}
\label{eq:subdiv_prob_update}
    q_i = \max{(\lambda_u \cdot q_{i}, \max{(\{b_0,...,b_n\})} - \min{(\{b_0,...,b_n\})})},
\end{equation}
where $\lambda_u$ and $\{b_0,...,b_n\}$ are the same as in \cref{eq:weight_update}.
In all experiments, we subdivide leaves every 500 iterations during the optimization if $q_i$ is above a threshold $\tau_s=0.5$.
Created leaves inherit their parents' probability and the initial value of $q_i$ is set to 0.
Note that, due to limitations of efficient multinomial sampling (see \cref{ssec:point_sampling_strategies}) with respect to numerical precision, we only subdivide if the resulting number of leaves is less than $256^3$.

\paragraph{Pruning.}
By repeatedly applying \cref{eq:weight_update}, the probability of leaf nodes whose volume is empty will exponentially decay towards zero.
As such we remove leaf nodes whose probability is less than $\tau_p=0.01$ every 100 iterations.
For stability, we only start pruning after the first 500 iterations.

\subsection{Point Sampling Strategies}
\label{ssec:point_sampling_strategies}
Fundamental to this work is the idea of sampling high-quality point clouds using the implicit octree-based point position estimator introduced in \cref{ssec:sampling_octree}.
Especially during optimization, we face the common issue of being limited in terms of GPU memory and are therefore restricted to a fixed budget for the number of sampled points.

This leads us to design two sampling strategies: A viewpoint-specific and a viewpoint-independent sampling scheme.
For \textit{training}, we want to generate a point cloud to render a specific viewpoint with which we can then compute the loss function in each iteration.
In contrast, a global, viewpoint-independent point cloud increases temporal stability as well as rendering performance during \textit{inference}, as no per-frame sampling of the implicit point cloud is required.
In the following, we detail the considerations that went into designing both sampling strategies.

\paragraph{Viewpoint-Specific Sampling.}
We identify three key properties of an effective re-weighting scheme for a specific viewpoint:
(1) No samples should be placed outside of the view frustum,
(2) regions further away from the camera require less samples,
and (3) leaves with a higher subdivision level $l$, i.e., those representing a smaller volume, should be sampled less.
Specifically, we compute the viewpoint-specific probability $\hat{p}_i$ as follows:
\begin{equation}
\label{eq:view_specific_reweighting}
\hat{p}_i=\mathds{1}_{visible} \cdot \frac{p_i}{d_i \cdot 2^{l_i \cdot \lambda_l}},
\end{equation}
where $\mathds{1}_{visible}$ is an indicator function returning one for visible leaves and zero otherwise, $\lambda_l$ controls how much less a smaller voxel should be sampled, and $d_i$ represents the distance between leaf center and the image plane. 
Empirically, we found $\lambda_l = 0.5$, i.e., a small bias towards sampling smaller voxels more often, to work well as it allows the model to more accurately reconstruct regions that are well-captured in training images.

Using multinomial sampling with replacement, we convert the re-scaled probabilities to a list of leaf indices.
For each element in this list, we uniformly sample the selected leaf's spatial extent to obtain the final 3D position.
This improves quality, as appearance features (see \cref{ssec:hash_grid_ar}) are regularized across a larger volume and not overwritten by hash-collisions.

For inference, we extend this scheme by sampling multiple point clouds, rasterizing them into 2D feature images (see \cref{ssec:differentiable_bilinear_point_splatting}), and averaging the features for each pixel.
Empirically, we find this to be superior to simply increasing the number of points for inference, which can be explained by the transmittance-based rendering formulation used by our rasterizer.

\paragraph{Viewpoint-Independent Sampling.}
Motivated by the observation that sampling an implicit point cloud takes up a large chunk of the rendering time during inference, we pre-extract a global point cloud for all viewpoints.
For this, we use \cref{eq:view_specific_reweighting} but omit the factors $\mathds{1}_{visible}$ and $d_i$.
Again, using multinomial sampling with replacement, we extract the number of samples for each leaf.
To increase stability, we then use the 3D Halton sequence~\cite{halton1964} for determining the final position of each sample.
We consider this as an advantage of our method: Our implicit formulation can be used to extract large point clouds without having to store them on disk.

\subsection{Appearance Representation}
\label{ssec:hash_grid_ar}
After sampling, we retrieve $M+1$ appearance features from a multi-resolution hash grid~\cite{mueller2022instant} for each point.
Before querying with each point's position, we apply the spherical contraction by \citet{barron2022mipnerf360} to increase the relative capacity of our model near the center of the scene, i.e., the best-observed region during optimization.
Of the retrieved per-point features, the first one is converted to a valid opacity value $\alpha_h \in [0, 1]$ using a Sigmoid-like activation: $\alpha_h = 1 - e^{-e^{x}}$~\cite{mueller2022instant}.
The remaining $M$ features are used as coefficients for SH evaluations to produce a view-dependent feature for each point.
In practice, we use SH of degree 2 and 4D view-dependent output features, resulting in $M=36$ SH coefficients per point.
As our background model, we employ a small MLP that computes a 4D feature for each pixel using the corresponding SH-encoded viewing direction as input.

\begin{table*}[tb]
\centering
\setlength\tabcolsep{14pt}
\scriptsize
\begin{tabular}{@{}lcccccc|ccc}
\toprule
                                      & \multicolumn{3}{c}{Mip-NeRF360}                                        & \multicolumn{3}{c|}{Tanks\&Temples}                                   & Train           & Render        & Size           \\
Method                                & LPIPS$\downarrow$      & SSIM$\uparrow$        & PSNR$\uparrow$        & LPIPS$\downarrow$     & SSIM$\uparrow$        & PSNR$\uparrow$        & hrs$\downarrow$ & fps$\uparrow$ & GB$\downarrow$ \\ \midrule
Instant-NGP~\cite{mueller2022instant} & 0.380                  & 0.698                 & 25.61                 & 0.438                 & 0.737                 & 21.82                 & 0.08            & 5.7           & 0.1            \\
ADOP~\cite{ruckert2022adop}           & 0.259                  & 0.723                 & 23.54                 & 0.236                 & 0.802                 & 21.69                 & 7.00            & 67.6          & 0.53           \\
TRIPS~\cite{franke2024trips}          & \cellcolor{zwei}0.213  & 0.778                 & \cellcolor{drei}25.94 & \cellcolor{zwei}0.229 & \cellcolor{drei}0.831 & 22.62                 & 4.00            & 90.0          & 0.76           \\
3DGS~\cite{kerbl3Dgaussians}          & 0.254                  & \cellcolor{drei}0.814 & \cellcolor{zwei}27.20 & 0.276                 & \cellcolor{zwei}0.866 & \cellcolor{drei}25.27 & 0.50            & 194.2         & 0.58           \\
Zip-NeRF~\cite{barron2023ICCV}        & \cellcolor{drei}0.219  & \cellcolor{zwei}0.828 & \cellcolor{eins}28.56 & \cellcolor{drei}0.233 & \cellcolor{eins}0.878 & \cellcolor{eins}26.75 & 5.00            & 0.2           & 0.9            \\
Ours                                  & \cellcolor{eins}0.164  & \cellcolor{eins}0.847 & \cellcolor{eins}28.56 & \cellcolor{eins}0.189 & \cellcolor{eins}0.878 & \cellcolor{zwei}25.93 & 9.58            & 2.9           & 1.1            \\ \midrule
Ours (16M)                            & 0.173                  & 0.841                 & 28.36                 & 0.220                 & 0.862                 & 25.12                 & 6.15            & 5.5           & 1.1            \\
Ours (8M)                             & 0.188                  & 0.830                 & 27.83                 & 0.252                 & 0.846                 & 24.54                 & 4.25            & 9.4           & 1.1            \\
Ours (pre-ex.)                        & 0.207                  & 0.802                 & 26.85                 & 0.261                 & 0.833                 & 23.71                 & 4.25            & 27.7          & 1.1            \\
\bottomrule
\end{tabular}
\caption{%
\textbf{Quantitative comparisons} on the Mip-NeRF360~\cite{barron2022mipnerf360} and Tanks and Temples~\cite{Knapitsch2017} datasets.
The three best results are highlighted in \textcolor{einsText}{\textbf{green}} in descending order of saturation.
Alongside our default configuration that uses 33M samples \textit{(Ours)} we also provide metrics for our method when trained with less samples (\textit{16M} and \textit{8M}). Furthermore, we report inference results using 67M points extracted with our view-independent sampling algorithm \textit{(pre-ex.)}.
}\label{tab:main_results}
\end{table*}

\subsection{Differentiable Bilinear Point Splatting}
\label{ssec:differentiable_bilinear_point_splatting}
Prior work on point rasterization demonstrated that point rendering for radiance fields can be very fast and yield great results~\cite{ruckert2022adop,kerbl3Dgaussians,franke2024trips}.
However, using one-pixel point rendering (projecting and discretizing points to one pixel) leads to aliasing as well as the need for approximate gradients~\cite{ruckert2022adop}.
To avoid this, we opt to use a bilinear formulation, that is we splat each point to the closest $2\times2$ pixels after projection. 
Thus, for a point $p_w = (x,y,z)^T$, we project it to the image coordinates $p = (u,v,d)^T$ with 
\begin{equation}
p = P \cdot V \cdot p_w,
\end{equation}
where $P$ and $V$ are the intrinsic and extrinsic camera matrices respectively.
For the $2\times2$ closest pixels' center points $p_{i\in\{0..3\}} = (u_i,v_i)^T$ we then compute the respective opacities with
\begin{equation}
\alpha = \alpha_h \cdot (1-|u-u_i|) \cdot (1-|v-v_i|).
\end{equation}
This causes the point's contributions to be weighted correctly based on its projected position.
We then use this while blending (see \cref{eq:alpha_blend_points}) all points $\mathcal{N}$ in depth order to render the image.

This bilinear splatting approach has three advantages: (1) We obtain more robust gradients, (2) improve temporal stability of the rendering pipeline, and (3) the rasterized images contain less holes which simplifies the task of the hole-filling CNN.
In TRIPS, \citet{franke2024trips} use a similar splatting approach but instead interpolate trilinearly into an image pyramid based on a learned per-point radius.
Our implicit point cloud enables us to render high-quality feature images without the need for interpolation with respect to a third dimension.

\subsection{Post-Processing}
\label{ssec:neural_pp_network}

For decoding, we use a standard three-layer U-Net architecture with a single residual block based on \textit{Fast Fourier Convolution (FFC)}~\cite{chi2020fast}.
We find that this enhances reconstruction regarding high-frequency details.
For challenging, e.g., auto-exposed outdoor scenes, we append the differentiable tone mapping module proposed by \citet{ruckert2022adop} to our pipeline.

\subsection{Optimization Loss}
\label{ssec:optimization_loss}
Inspired by prior works~\cite{kerbl3Dgaussians, hahlbohm2023plenopticpoints}, we combine a per-pixel loss and two established image-space loss functions:
\begin{equation}
\textstyle \mathcal{L} = \mathcal{L}_\text{R} + \mathcal{L}_\text{D-SSIM} + \lambda_\text{vgg} \cdot \mathcal{L}_\text{VGG} + \lambda_\text{decay} \cdot \mathcal{L}_\text{Reg}.
\end{equation}
Specifically, we use the robust loss $\mathcal{L_\text{R}}$~\cite{barron2019general} with $\alpha=0$ and $c=0.2$ as our per-pixel loss as well as D-SSIM and VGG~\cite{vgg16loss} losses which are commonly attributed with a closer resemblance of human perception.
For regularization we follow \citet{barron2023ICCV} and impose a normalized weight decay on the parameters of the multi-resolution hash grid.
We use $\lambda_\text{vgg}=0.075$ and $\lambda_\text{decay}=0.1$.

\section{Experiments}

We conduct multiple experiments to evaluate our method.

\subsection{Datasets and Baselines}
\label{ssec:eval_implementation_datasets_baselines}
For evaluation, we use a total of 17 real scenes featuring a broad spectrum of challenges regarding both geometric and photometric aspects.
The Mip-NeRF360 dataset~\cite{barron2022mipnerf360} contains five outdoor and four indoor scenes captured with fixed exposure and white balance settings.
We further use all eight scenes from the \textit{intermediate} set of the Tanks and Temples dataset~\cite{Knapitsch2017}.
It was captured without fixed camera settings and presents challenges regarding photometric variation that complicate reconstruction, providing relevant insights for in-the-wild performance.
We use the established 7:1 train/test split~\cite{barron2022mipnerf360} for all scenes.
We optimize for $50,000$ iterations and render an image for a single viewpoint in each of those.
See \cref{sec:implementation} for further details regarding our implementation.

We compare our method against Instant-NGP~\cite{mueller2022instant}, ADOP~\cite{ruckert2022adop}, TRIPS~\cite{franke2024trips}, 3D Gaussian Splatting~\cite{kerbl3Dgaussians}, and the current state of the art in terms of image quality Zip-NeRF~\cite{barron2023ICCV}.
We use the images kindly provided by the authors of the respective publications when available -- otherwise we use the official implementation to generate the images -- and compute all image quality metrics under identical conditions.
All methods use a RTX 4090 when memory was sufficient, otherwise an A100 was used.

\subsection{Results}
We show quantitative results for the scenes from Mip-NeRF360 and Tanks and Temples in \cref{tab:main_results}, as well as qualitative comparisons in \cref{fig:teaser}. For per-scene metrics and additional visual comparisons see \cref{sec:per_scene_results}.
In terms of image quality metrics, our method clearly outperforms previous point-based techniques (TRIPS and 3DGS) and achieves similar quality as Zip-NeRF on both datasets.
Visually, we observe that our method outperforms all baselines with respect to representing fine details, which is also represented in its excellent LPIPS scores.
We complement our evaluation by conducting a perceptual experiment in which we compare INPC against Zip-NeRF, as the latter achieves the best quality metrics among the compared-against methods (see \cref{tab:main_results}).
We followed a fully randomized, within-participants experimental design with a 2AFC task.
Our $17$ participants saw the results of both methods side-by-side (one pair at a time, in random order and screen side, with a different order per participant) and were instructed to select the image they preferred.
The $55$ stimuli covered all 17 evaluated scenes and consisted of a minimum of $3$ frames per scene.
Our method was favored by the participants on an average of $69.41\%$ of the cases, with all participants preferring our results with a ratio above the chance line. 
See \cref{sec:experiment} for full details on the experimental setup, stimuli selection, and evaluation.

We further show approximates for the training time, inference frame rate, and resulting model size in \cref{tab:main_results}.
Our method requires slightly longer training than the recent Zip-NeRF and TRIPS.
Like other NeRF-based methods such as Zip-NeRF, our model always has the same size (1.1 GB), whereas point-based methods such as 3DGS require up to 2 GB of storage depending on the scene.
Regarding inference frame rate, our method is roughly an order of magnitude faster than Zip-NeRF but currently outperformed by the explicit point-based approaches 3DGS, TRIPS, and ADOP.

In \cref{tab:render_profiling}, we break down inference performance for different versions of our model that uses 8M samples during training.
While this work focuses on image quality, the numbers confirm that even at FHD resolution, our model maintains its interactivity.

\begin{table}[tb]
\centering
\setlength\tabcolsep{6.3pt}
\scriptsize
\begin{tabular}{@{}lccc|cc}
\toprule
Method              & Sampling & Rendering & Post-Proc. & Total   & \#Points     \\
\midrule
Ours (8M)           & 51ms     & 37ms      & 26ms       & 114ms   & 4$\times$8M  \\
Ours (8M)$^\dagger$ & 16ms     & 10ms      & 26ms       & 52ms    & 8M           \\
Ours (pre-ex.)      & N/a      & 28ms      & 26ms       & 54ms    & 67M          \\
\bottomrule
\end{tabular}
\caption{%
\textbf{Inference speed breakdown} for the \textit{Playground} scene~\cite{Knapitsch2017} rendered on an RTX 4090 GPU (2000$\times$1085 pixels). The configuration marked with $\dagger$ is the same as in ablation F) shown in \cref{tab:ablation_tables_tab}. The bottleneck of our rendering is the sorting step which requires 27 / 7 / 17 milliseconds (ms) respectively.
}\label{tab:render_profiling}
\end{table}

\paragraph{Ablations.}

\begin{table}[b!]
\centering
\setlength\tabcolsep{13.5pt}
\scriptsize
\begin{tabular}{@{}lccc}
\toprule
Configuration                     & LPIPS$\downarrow$     & SSIM$\uparrow$        & PSNR$\uparrow$        \\ 
\midrule
\tabbox{A)} No D-SSIM Loss        & 0.204                 & 0.729                 & 25.27                 \\
\tabbox{B)} No VGG Loss           & 0.238                 & 0.754                 & 25.29                 \\
\tabbox{C)} No Weight Decay       & 0.210                 & 0.744                 & 25.22                 \\
\tabbox{D)} No Subdivision        & 0.508                 & 0.406                 & 19.15                 \\
\tabbox{E)} No Bilinear Splatting & 0.243                 & 0.708                 & 24.36                 \\
\tabbox{F)} No Multisampling      & 0.201                 & 0.740                 & 25.02                 \\
\tabbox{G)} No SfM Prior          & 0.197                 & 0.753                 & 25.29                 \\
\tabbox{H)} No Background Model   & 0.197                 & 0.752                 & 25.28                 \\
\tabbox{I)} No FFC Block          & 0.207                 & 0.749                 & 25.32                 \\ 
\tabbox{J)} No Post-Processing    & 0.277                 & 0.718                 & 24.32                 \\
\midrule
Ours (8M)                         & 0.192                 & 0.761                 & 25.53                 \\
\bottomrule
\end{tabular}
\caption{%
\textbf{Model ablations} with respect to image quality metrics computed on the outdoor scenes from the Mip-NeRF360 dataset.
}\label{tab:ablation_tables_tab}
\end{table}

\begin{figure}[hb]
\centering
\includegraphics[width=\linewidth, keepaspectratio]{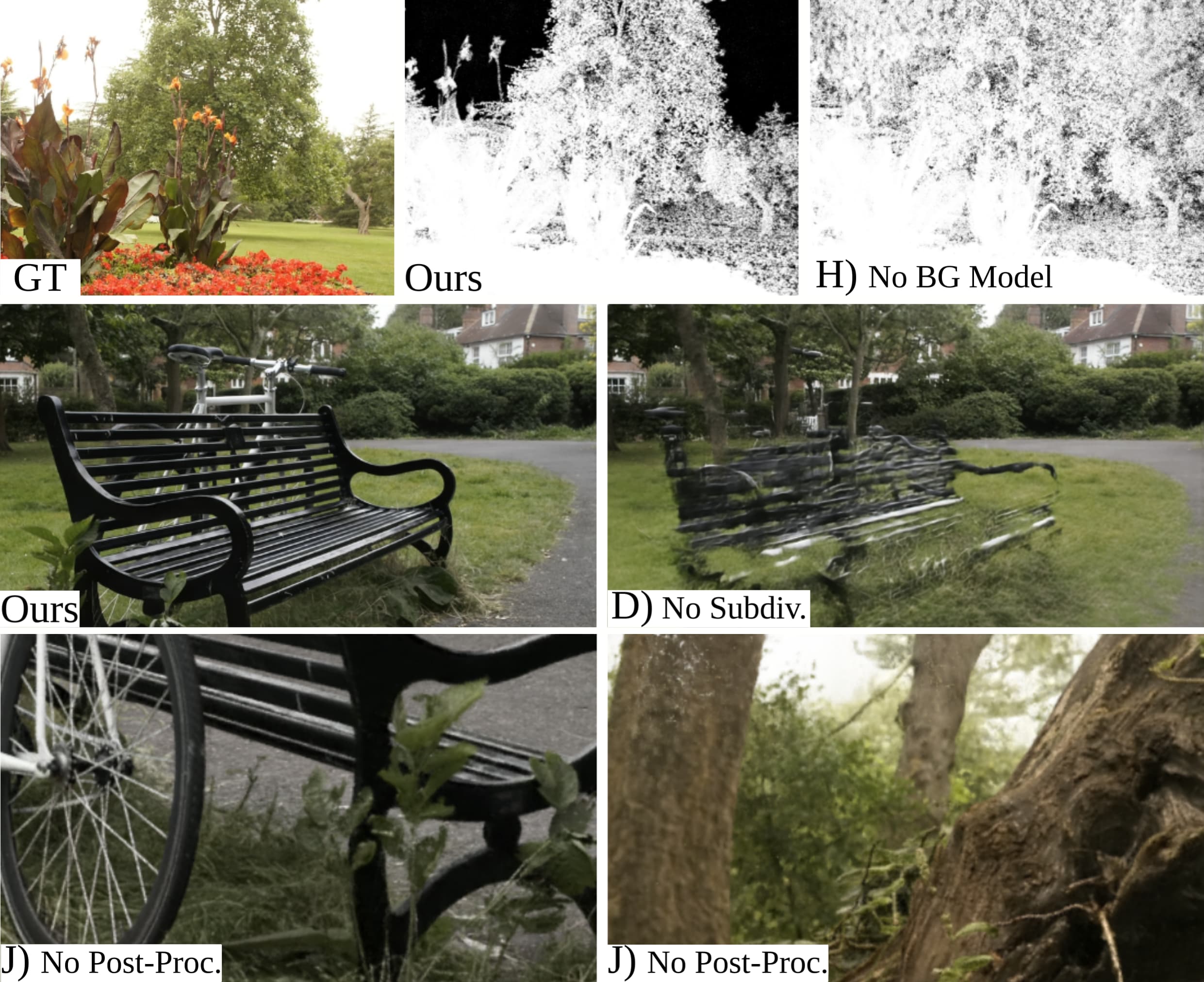}
\caption{%
\textbf{Visual comparisons} for ablations H, D, and J 
in \cref{tab:ablation_tables_tab}.
Our background model prevents the sampling of points in the sky. Disabling octree subdivision causes foreground reconstruction to fail.
Omitting post-processing (\cref{ssec:neural_pp_network}) leads to holes and high-frequency noise in renderings.
}\label{fig:ablation_tables_images}
\end{figure}

In \cref{tab:ablation_tables_tab} and \cref{fig:ablation_tables_images}, we show ablations of our 8M model computed on the five outdoor scenes from the Mip-NeRF360 dataset.
A)~--~C) dissect our loss function showing that each term has a meaningful contribution.
For D), we omit octree subdivision during optimization, which greatly hinders reconstruction of the scene's foreground while having no visible impact on far away objects.
E) and F) validate the effectiveness of our bilinear splatting approach as well as the point cloud multisampling during inference.
Next, G) shows that our method does not depend on initial SfM points for sampling probability initialization to achieve its high visual fidelity.
For H), we observe that leaving out the background model barely impacts metrics but results in sampling points in the sky (see alpha images in \cref{fig:ablation_tables_images}) which results in "floaters" upon visual inspection.
Furthermore, I) indicates the residual FFC block in the U-Net enhances performance for high-frequency details, which is only partly captured by metrics.
Lastly, we omit our post-processing, i.e., the U-Net and tone mapping module described in \cref{ssec:neural_pp_network}, in J).
As this leaves our point-based method with no hole-filling technique, we double the number of samples during training to 16M and also disable $\lambda_\text{vgg}$ due to the absence of the CNN.
Somewhat surprisingly, we observe both quantitatively and qualitatively (see \cref{fig:ablation_tables_images}) that this configuration still provides good results, which confirms the potential of our implicit point cloud formulation.

\begin{figure*}[tbh]
\centering
\includegraphics[width=\linewidth, keepaspectratio]{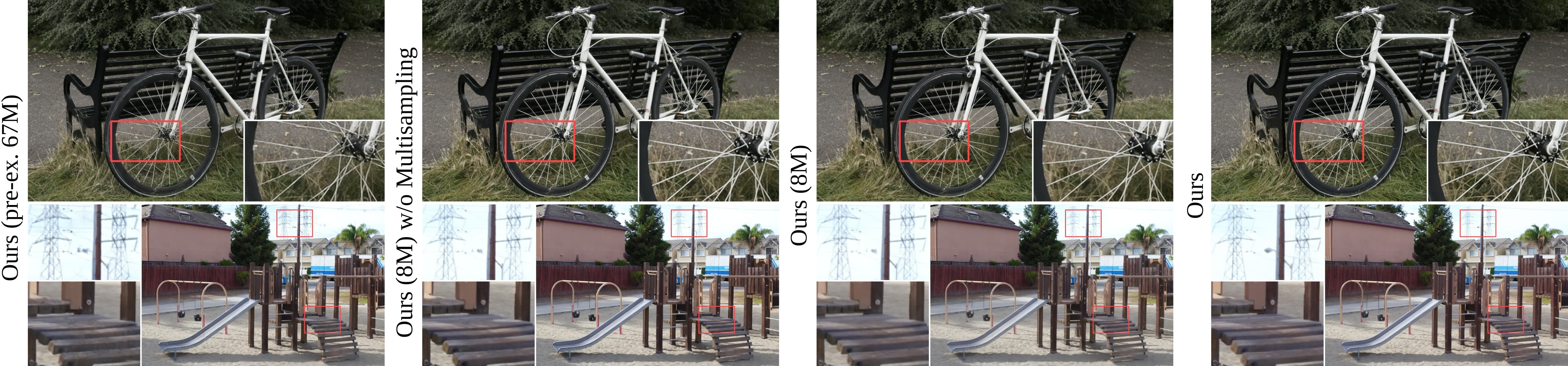}
\caption{%
\textbf{Visual comparison} of INPC configurations. Our global pre-extraction slightly reduces visual quality in terms of fine detail, especially in FHD renderings of Tanks and Temples scenes. Without multisampling images are slightly sharper but our sampling sometimes misses thin structures. For Mip-NeRF360 scenes, the difference between our \textit{8M} and default (\textit{33M}) configuration is barely visible.
}\label{fig:model_config_comparisons}
\end{figure*}

\section{Discussion}
The experiments confirm our approach's effectiveness on common benchmarks, outperforming current state of the art regarding perceived image quality, while rendering at interactive frame rates.
This is also reflected in the per-scene image quality metrics (see \cref{sec:per_scene_results}).

All our models use $4\times$ multisampling for inference, which results in up to 528 million blended splats (4$\times$33M points) per rendered feature image for our default configuration that uses 33M samples in each training iteration.
We would like to highlight the results of our 8M configuration.
It provides excellent quality, while training in just over 4 hours and blending 128 million splats (4$\times$8M points) into each feature image at interactive frame rates of $\sim$9 fps -- all on a single RTX 4090 GPU.
Albeit at the cost of image quality, it allows for easy modification towards faster rendering by either disabling multisampling or by pre-extracting a global point cloud.
Moreover, our view-independent sampling algorithm causes only a small drop in visual quality for the Mip-NeRF360 dataset, as evident by the LPIPS metric, where it still outperforms 3DGS.
However, we observe that due to the higher image resolution (roughly $2\times$ more pixels) of the Tanks and Temples scenes, the budget of 67M pre-extracted global points is not always sufficient.
We show visual comparisons for our default, 8M, and pre-extracted configuration in \cref{fig:model_config_comparisons}.

\begin{figure}[b!]
\centering
\includegraphics[width=\columnwidth, keepaspectratio]{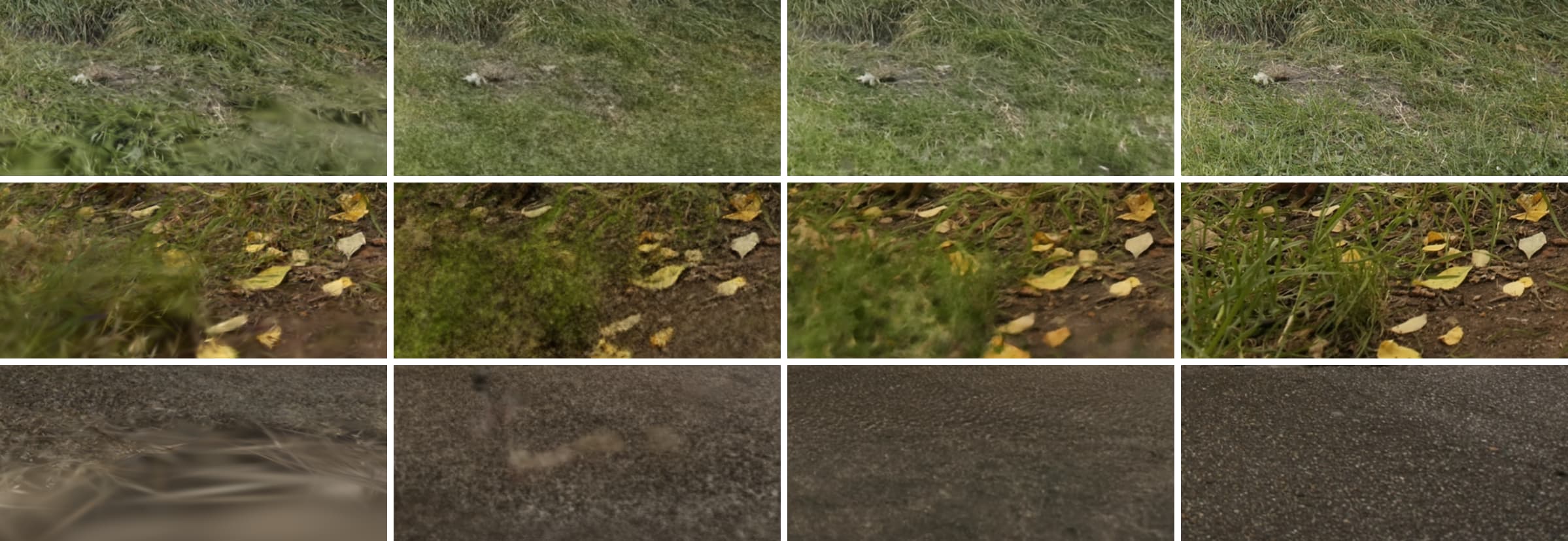}
\vspace{-7pt}
\begin{tabular}{*{4}{p{0.2\columnwidth}<{\centering}}}
\vspace{-13pt} \scriptsize 3DGS~\cite{kerbl3Dgaussians} & \vspace{-13pt}\scriptsize Zip-NeRF~\cite{barron2023ICCV} &  \vspace{-13pt}\scriptsize Ours &  \vspace{-13pt}\scriptsize Ground Truth
\end{tabular}
\caption{%
\textbf{Limitations.} Our method and state-of-the-art baselines sometimes fail to recover fine geometric detail near the camera.
}\label{fig:limitations}
\end{figure}

\paragraph{Limitations.}
Regarding limitations of our method, we observe that it is sometimes unable to reconstruct fine geometric detail close to the camera, a property shared with existing methods such as 3DGS and Zip-NeRF (see \cref{fig:limitations}).
Similar to previous works that used a CNN for post-processing rasterizer outputs~\cite{ruckert2022adop, franke2023vet}, we identify temporal stability as a minor issue during inference (see our supplemental video).
While this work is mostly focused on image quality, we observe that the global sorting used in our implementation impacts its real-time rendering capability.
This is a disadvantage compared to the recent 3DGS and TRIPS.
However, we are confident that an optimized implementation can overcome some of the gap, as specialized rendering methods for explicit point clouds showed promising results~\cite{schutz2022software}.

As extracting a global point cloud greatly boosts frame rate during inference, the optimization pipeline could be adjusted to facilitate viewpoint-independent sampling, e.g., by encouraging binary opacity values as done by \citet{reiser2024binary}.
We also believe the underlying octree-based data structure could further be improved.
Lifting our implementation's limitation of $256^3$ active leaf nodes in the octree, in combination with improved routines for subdivision, updating, and pruning, is likely to boost reconstruction quality.

\section{Conclusion}
In this work, we have introduced Implicit Neural Point Clouds, a concept fusing NeRF- and point-based radiance fields, utilizing the advantages of both.
Our INPC retains favorable optimization properties of NeRF by representing a point cloud inside an octree-based probability field for point positions and an implicit appearance model.
The evaluation shows that our method improves upon the previous state-of-the-art method in terms of perceptual image quality, while also enabling rendering at interactive frame rates on consumer-grade hardware.
We believe that the implicit point cloud representation as well as other ideas presented here can enable future work towards further closing the gap between best-quality and real-time radiance field approaches.

\begin{center}
\url{https://fhahlbohm.github.io/inpc/}
\end{center}

\paragraph{Acknowledgments.} 
We would like to thank Peter Kramer for his help with the video, Timon Scholz for his help with the implementation of our viewer, and Fabian Friederichs and Leon Overkämping for their valuable suggestions.
This work was partially funded by the DFG (“Real-Action VR”, ID 523421583) and the L3S Research Center, Hanover, Germany.
We thank the Erlangen National High Performance Computing Center (NHR@FAU) for the provided scientific support and HPC resources under the NHR project b162dc. NHR funding is provided by federal and Bavarian state authorities. NHR@FAU hardware is partially funded by the DFG (ID 440719683).
Linus Franke was supported by the Bavarian Research Foundation (AZ-1422-20) and the 5G innovation program of the German Federal Ministry for Digital and Transport under the funding code 165GU103B.

{
    \small
    \bibliographystyle{ieeenat_fullname}
    \bibliography{references}
}

\appendix
\section{Implementation}
\label{sec:implementation}
We believe that the flexibility of our implicit point cloud could prove to be useful for future work. Here, we try to support such work by providing all details of our implementation.

\subsection{Overview}
We implement our method using PyTorch and create re-usable PyTorch extensions using C$++$$\slash$CUDA for our differentiable rasterization module and multiple -- otherwise slightly slower -- stages of our sampling process.
Furthermore, we use multiple libraries that provide efficient implementations for parts of our pipeline.
For the multi-resolution hash grid as well as the background MLP, we use the implementation provided with the \textit{tiny-cuda-nn (TCNN)} framework~\cite{tiny-cuda-nn}.
As our optimizer, we use the fused ADAM~\cite{kingma2014adam} implementation provided by the NVIDIA Apex library.
To reduce computation time of the CNN, we also make use of PyTorch's automatic mixed precision package and scale FP16 gradients by a fixed factor of $128$ during the backward pass.

\subsection{Pre-Processing Details}
\myparagraph{Lens Distortion.}
For the Tanks and Temples dataset~\cite{Knapitsch2017}, the original images are subject to slight lens distortions.
While our rendering supports radial and tangential distortion, we use COLMAP~\cite{schoenberger2016colmap1, schoenberger2016colmap2} to extract undistorted images after camera calibration to ensure fair comparison against the selected baselines.
Specifically, 3DGS~\cite{kerbl3Dgaussians} only works with a pinhole camera model.
Note that the undistorted images are used by all of the compared methods, including INPC.

\myparagraph{Scene Normalization.}
We follow \citet{barron2022mipnerf360} and apply a world-space transformation to all camera poses so that they fit inside a cube $[-1, 1]^3$.
We apply the same transformation to the SfM point cloud if it is used.
For all scenes, we set the values for near and far plane to $0.01$ and $100$ respectively.

\subsection{Optimization Details}
For ADAM, we use $\beta_1=0.9$, $\beta_2=0.99$, $\epsilon=10^{-15}$, and disable weight decay.
The learning rates for all parameters are exponentially decayed during the optimization.
For the hash grid and MLP parameters, we decay the learning rate from $1$e$-2$ to $3$e$-4$, while CNN parameter learning rates are decayed from $3$e$-4$ to $5$e$-5$.

When using differentiable tone mapping as proposed by \citet{ruckert2022adop}, we follow the authors and only activate the sub-modules for exposure and camera response.
Similar to \citet{ruckert2022adop}, these parameters are updated with a fixed learning rate of $5$e$-4$ and $1$e$-3$ respectively.
However, we use a warm-up strategy~\cite{barron2021mipnerf} where the learning rate cosine-delayed by multiplying with a factor between $0.01$ and $1$ during the first $5,000$ iterations.
This is slightly different from the original version that simply disabled optimization of tone mapping parameters during the first few iterations.

\subsection{Point Probability Octree Details}
\label{ssec:appendix_octree_details}
\myparagraph{Initialization.}
For each scene, we define $b_{\min}, b_{\max} \in \mathbb{R}^3$ as the axis-aligned bounding box of the SfM point cloud and enlarge the result by $10$\% for increased robustness.
If no SfM point cloud is available, we instead use the smallest axis-aligned box enclosing the viewing frustra of all training cameras.
Given a desired initial resolution $R$ and the aforementioned bounding box of the scene, we initialize our octree-based data structure as a regular grid of cube-shaped voxels.
These 3D voxels are the initial leaf nodes of $\mathcal{P}$.
For each leaf node, we store its center $c_i\in\mathbb{R}^3$ in world space, current subdivision level $l_i\in\mathbb{N}_0$, sampling probability $p_i\in[0, 1]$, and subdivision "probability" $q_i\in[0, 1]$.
We limit values of $R$ to powers of $2$ and can thus initialize the initial subdivision level as $\log_2(N)$ for all leaf nodes.
To incorporate a point cloud $\mathcal{I}$ as a prior, we modify initial sampling probability:
Assuming $|\mathcal{I}_i|$ is the number of points inside the volume of the $i$-th leaf node and $Q$ is the $0.95$-quantile of these point counts, we compute the initial probability as $\frac{|\mathcal{I}_i|}{Q}$, clipping values with $0.1$ and $1.0$ for increased robustness.

\myparagraph{Ensuring Sample Count.}
For our viewpoint-specific sampling, a major difficulty comes from dealing with those leaves whose volume intersects the faces of the viewing frustum.
For these, it is possible that positions generated by our sampling scheme lie outside the frustum.
To this end, we re-sample until the desired number of samples is reached.
Note that our re-weighting scheme for sampling probabilities does not account for partial visibility of voxels.
This means simply placing points inside the visible volume of each leaf would slightly bias sampling towards partially visible leaves.
We avoid this by re-sampling the viewpoint-specific probability distribution for affected samples repeatedly until all generated samples are visible.
Notably, our multisampling approach allows us to omit this step during inference, which speeds up sampling. 

\myparagraph{Depth Calculation.}
For our view-specific re-weighting scheme, we look to obtain a view-specific depth value $d_i \in [0, 1]$ for each leaf node.
To this end, we adjust the distance of each leaf center to the image plane, i.e., the view-space depth value $z_{\text{view}}$:
\begin{equation}
    d_i=\max{(\frac{|z_{\text{view}} - \text{near}|}{\text{far}}, \tau_d)}.
\end{equation}
Note that $|z_{\text{view}} - \text{near}|$ accounts for the fact that $z_{\text{view}}$ may be smaller than "near" for partially visible leaves (see the previous paragraph).
To prevent extreme cases of oversampling close to the camera, we use a threshold $\tau_d=10^{-8}$ in all experiments.

\subsection{TCNN Details}
For the multi-resolution hash grid and the accompanying MLP, we use the implementation from the TCNN framework~\cite{tiny-cuda-nn}.
We tested different configurations using those employed by prior works~\cite{mueller2022instant,yariv2023bakedsdf,barron2023ICCV} as a starting point.
Our resulting configuration is shown in \cref{lst:hash_grid_config}.
It is similar to the one used by Zip-NeRF~\cite{barron2023ICCV}, but we increase the hashmap size from $2^{21}$ to $2^{23}$ (see \cref{sec:additional_ablations} for ablations).
For the background MLP we also use the fully-fused implementation from the TCNN framework~\cite{tiny-cuda-nn}.
It has four hidden layers with $128$ neurons each and uses ReLU activations.
For each pixel, we encode its normalized viewing direction using spherical harmonics of degree three before inputting into the MLP.
Note that in TCNN this means using \verb|'degree': 4| and converting input directions to $[0, 1]^3$.
\begin{listing}[b!]
\caption{\label{lst:hash_grid_config}
\textbf{Hash grid configuration} for the \textit{tiny-cuda-nn} implementation~\cite{tiny-cuda-nn}.
}
\vspace{-7pt}
\begin{minted}[frame=none,fontsize=\scriptsize,framesep=0mm,baselinestretch=0.5]{python}
encoding_config = {
    'otype': 'Grid',
    'type': 'Hash',
    'n_levels': 10,
    'n_features_per_level': 4,
    'log2_hashmap_size': 23,
    'base_resolution': 16,
    'per_level_scale': 2.0,
    'interpolation': 'Linear'
}
network_config = {
    'otype': 'FullyFusedMLP',
    'activation': 'ReLU',
    'output_activation': 'None',
    'n_neurons': 64,
    'n_hidden_layers': 1
}
\end{minted}
\vspace{-7pt}
\end{listing}

\subsection{Bilinear Splatting Details}
Prior work on point rasterization has demonstrated that software implementations are a) very fast~\cite{SCHUETZ-2021-PCC, schutz2022software} compared to hardware implementations, e.g., \verb|GL_POINTS|, and b) allow for backward pass implementations suited for gradient-based optimization~\cite{lassner2021pulsar, ruckert2022adop}.
Motivated by these insights, we set out to design our own rasterization module in an effort to adapt it to the needs of our rendering pipeline.

Part of why MLP-based radiance fields works so well is the fully differentiable color accumulation of multiple samples for a given pixel.
While NeRF uses a volume rendering-based formulation, recent successful point-based methods~\cite{kerbl3Dgaussians, franke2024trips} use conventional $\alpha$-blending, which is similarly favorable for gradient-based optimization.
We deem the extra cost of sorting points worthy with respect to to the superior optimization properties.

The second consideration is the selection of points blended into each pixel.
Typically, each point is projected onto a single pixel using the extrinsic and intrinsic camera matrices.
Using this approach, the unstructured and disconnected nature of point clouds may cause large holes in the resulting image.
To this end, we follow the common practice of optimizing a hole-filling CNN alongside the rest of our model, which helps a great deal but causes other downstream problems such as temporal instabilities.
However, other approaches for addressing the issue exist.
Among these, splatting is a well-established technique in rasterization-based rendering where the same point may influence multiple pixels at once.
Furthermore, splatting ameliorates the problems arising from the need of discrete pixel coordinates for drawing each point.
While, mathematically speaking, a point is always projected onto a single pixel, giving the same weight to points projected near pixel boundaries and those projected into the center of a pixel can be problematic for gradient-based optimization.
We therefore splat each point into the four closest pixels and achieve this by downweighting its opacity using bilinear interpolation before blending it into each pixel (see main paper).
During $\alpha$-blending, we stop accumulation for a pixel once the remaining transmittance falls below $10^{-4}$.
The full algorithm, which we implement in CUDA to achieve accelerated rendering times, can be written as shown in \cref{alg:rasterize}.
\begin{algorithm}[b!]
\caption{GPU rasterization with bilinear splatting and $\alpha$-blending}
\label{alg:rasterize}
\begin{scriptsize}
\begin{algorithmic}[1]
\Require $C$: Camera model
\Require $P, A, F$: Point positions, opacities, and features
\Ensure $I$: Image
\Function{Rasterize}{$C, P, A, F$}
    \State $P' \gets \textsc{ProjectAndCull}(P, C)$
    \State $A', K, N \gets \textsc{BilinearSplatting}(P', A)$ \Comment{Splat $\alpha$, keys, counts}
    \State $L \gets \textsc{ExclusiveSum}(N)$ \Comment{Splat indices}
    \State \textsc{RadixSortPairs}($K, L$) \Comment{Global sort}
    \State $I \gets \textsc{BlendSortedSplats}(A', F, L, N)$ \Comment{Per-pixel $\alpha$-blending}
    \State \Return $I$
\EndFunction
\end{algorithmic}
\end{scriptsize}
\end{algorithm}
We use the NVIDIA CUB library for the exclusive sum as well as the global radix sorting of key$\slash$value pairs.
As we identify global memory accesses to be the main performance bottleneck for our blending kernel, we implement it using CUDA's warp-level primitives distributing the workload for each pixel over a full warp of $32$ threads.
The blending weight computation required for updating $\mathcal{P}$ as well as the spherical harmonics computations are fused into this pipeline.
During optimization, we store the bilinear interpolation weights, per-pixel point counts, and sorted key$\slash$index buffers for the backward pass.
To compute per-splat gradients we process them in the same order as in the forward pass.
We obtain a per-point gradient by combining its four splat gradients in a weighted sum according to the respective bilinear interpolation weights, i.e., the analytical derivative of the bilinear splatting process.
For the probability field updates, we also need to combine per-splat blending weights of each point.
Interestingly, we observe that the use of a weighted sum according to the respective bilinear interpolation weights outperforms the theoretically correct approach of a simple sum in all our experiments.
We believe this is because the weighted sum causes the resulting per-point blending weight to more closely resemble the expected visibility of future samples from the leaf from which the point was sampled.

\subsection{CNN Details}
For the rendering network, we use a standard three-layer U-Net architecture with $64$ initial filters, GELU activations, average pooling for downsampling, and bilinear interpolation followed by a point-wise convolution for upsampling.
One important change is the introduction of a single residual block based on \textit{Fast Fourier Convolution (FFC)}~\cite{chi2020fast}.
We use the authors' implementation with $\alpha_{in},\alpha_{out} = 0.75$ and no further modifications.
To avoid having to crop or pad when concatenating tensors for the employed skip-connections in the expansive path of the U-Net, we pad the target width and height to a multiple of $2^{L - 1}$, where $L$ is the number of layers in the U-Net, before initiating our rendering pipeline.
Note that this requires adjusting the principal point to allow for renderings that -- after removing the padded pixels -- can properly be compared to ground truth images for loss or quality metric computation.

\begin{table}[thbp]
\centering
\setlength\tabcolsep{3.18pt}
{\scriptsize
\begin{tabular}{@{}lcccccc@{}}
\toprule
\multirow{2}{*}{Configuration}                          & \multirow{2}{*}{LPIPS$\downarrow$} & \multirow{2}{*}{SSIM$\uparrow$} & \multirow{2}{*}{PSNR$\uparrow$} & Train & Render& Size\\
                        &  & & & (hrs)$\downarrow$ & (fps)$\uparrow$ & (GB)$\downarrow$ \\
\midrule
\tabbox{A)} No D-SSIM Loss             & 0.204             & 0.729          & 25.27          & 4.24                       & 9.8                       & 1.1                         \\
\tabbox{B)} No VGG Loss                & 0.238             & 0.754          & 25.29          & 3.43                       & 9.8                       & 1.1                         \\
\tabbox{C)} No Weight Decay            & 0.210             & 0.744          & 25.22          & 3.69                       & 9.8                       & 1.1                         \\
\tabbox{D)} No Subdivision             & 0.508             & 0.406          & 19.15          & 4.07                       & 9.9                       & 0.88                        \\
\tabbox{E)} No Bilinear Splatting      & 0.243             & 0.708          & 24.36          & 4.14                       & 11.7                      & 1.1                         \\
\tabbox{F)} No Multisampling           & 0.201             & 0.740          & 25.02          & 4.25                       & 26.3                      & 1.1                         \\
\tabbox{G)} No SfM Prior               & 0.197             & 0.753          & 25.29          & 4.25                       & 9.8                       & 1.1                         \\
\tabbox{H)} No Background Model        & 0.197             & 0.752          & 25.28          & 4.24                       & 9.9                       & 1.1                         \\
\tabbox{I)} No FFC Block               & 0.207             & 0.749          & 25.32          & 4.15                       & 10.0                      & 1.1                         \\
\tabbox{J)} No Post-Processing         & 0.277             & 0.718          & 24.32          & 4.05                       & 5.6                       & 1.1                         \\
\midrule
$\mathcal{P}$: No Viewpoint Bias       & 0.328             & 0.618          & 23.23          & 3.83                       & 10.1                      & 1.0                         \\
$\mathcal{P}$: No Depth Re-weighting   & 0.226             & 0.721          & 24.72          & 4.25                       & 9.8                       & 1.1                         \\
$\mathcal{P}$: $d_i \rightarrow d_i^2$ & 0.210             & 0.740          & 25.04          & 4.25                       & 9.8                       & 1.1                         \\
$\mathcal{P}$: No Size Re-weighting    & 0.198             & 0.753          & 25.27          & 4.25                       & 9.8                       & 1.1                         \\
$\mathcal{P}$: $\lambda_l=2$           & 0.363             & 0.557          & 22.01          & 4.25                       & 9.8                       & 1.1                         \\
$\mathcal{A}$: No Space Contraction    & 0.242             & 0.702          & 24.59          & 4.45                       & 10.2                      & 1.1                         \\
$\mathcal{A}$: Hashmap Size $2^{21}$   & 0.203             & 0.750          & 25.23          & 3.57                       & 10.6                      & 0.52                        \\
$\mathcal{A}$: Hashmap Size $2^{22}$   & 0.201             & 0.751          & 25.34          & 3.76                       & 10.4                      & 0.73                        \\
$\mathcal{A}$: No View-Dependence      & 0.205             & 0.743          & 24.80          & 4.13                       & 11.6                      & 1.1                         \\
$\mathcal{A}$: SH Degree 1             & 0.196             & 0.753          & 25.32          & 4.16                       & 10.4                      & 1.1                         \\
$\mathcal{A}$: 3 SG Lobes              & 0.197             & 0.753          & 25.26          & 4.58                       & 8.2                       & 1.1                         \\
$\mathcal{A}$: 4 SG Lobes              & 0.197             & 0.755          & 25.26          & 4.85                       & 7.8                       & 1.1                         \\
$\mathcal{A}$: 1 ASG Lobe              & 0.200             & 0.751          & 25.16          & 4.69                       & 8.4                       & 1.1                         \\
U-Net: 32 Initial Filters              & 0.206             & 0.750          & 25.23          & 3.84                       & 10.6                      & 1.1                         \\
U-Net: 2 Layers                        & 0.207             & 0.745          & 24.88          & 4.23                       & 9.8                       & 1.1                         \\
Ours (4M)                              & 0.228             & 0.733          & 24.93          & 3.61                       & 17.5                      & 1.1                         \\
\midrule
Ours (8M)                              & 0.192             & 0.761          & 25.53          & 4.25                       & 9.8                       & 1.1                         \\
\bottomrule
\end{tabular}
}
\caption{%
Model ablations computed on the five Mip-NeRF360 outdoor scenes.
}\label{tab:extra_ablations}
\end{table}

\section{Additional Ablations}
\label{sec:additional_ablations}
In \cref{tab:extra_ablations}, we show additional model ablations computed by running various versions of our 8M configuration on the five outdoor scenes from the Mip-NeRF360 dataset~\cite{barron2022mipnerf360}.
For all configurations, we also include model size as well as optimization time and inference frame rate measured on the \textit{Bicycle} scene ($1237 \times 822$ pixels).
To avoid any confusion, we want to mention that we re-computed the results of the baseline configuration \textit{Ours (8M)} for the ablation studies.
As our method produces \textit{marginally} different results in every run, the results are not exactly the same as in \cref{tab:per_scene_metrics_m360}.
This also applies to the corresponding tables from the main paper.
The reason for these slightly different results lies in the fact that the PyTorch modules we use for the CNN use fast implementations provided by the CUDA Deep Neural Network (cuDNN) API.
For floating point numbers, applying associative operations in different order might slightly change the result.
Many cuDNN kernels do not ensure identical ordering of operations for every kernel launch to obtain faster computation times.
Therefore, the slight differences in results can not be prevented by, e.g., using a fixed random seed -- which we are doing anyways.

\myparagraph{Viewpoint-Specific Sampling.}
We show the impact of omitting$\slash$changing parts of our view-specific re-weighting scheme used when sampling our point probability octree $\mathcal{P}$.
Without any re-weighting, i.e., sampling the whole scene for every viewpoint, we observe a significant drop in quality of the reconstruction.
We further observe that the depth term $\nicefrac{1}{d_i}$ is more important than the size term $\nicefrac{1}{2^{l_i \cdot \lambda_l}}$.
While intuition may suggest that the depth term should be squared for a perspective camera model to not overemphasize on the background, we show that this leads to worse results.
This is caused by the use of a CNN for hole-filling, as it can easily fill slightly larger holes close to the camera but may not correctly produce clean background with less samples.
As a result, placing more samples in the background is beneficial for our method.
Along the same lines, we expected $\lambda_l = 2$ to work best, as it most closely resembles the relative size of leaves when projected onto the 2D image plane.
However, this performs much worse than using $\lambda_l = 0.5$, which we think is due to the fact that for better observed regions leaves are more likely to get subdivided.
Thus, overemphasizing smaller leaves leads to more samples being placed into well-reconstructed areas, which improves results.

\myparagraph{Appearance Field.}
Our appearance field $\mathcal{A}$ requires querying with positions in $[0, 1]^3$, for which a employ spherical contraction~\cite{barron2022mipnerf360}.
If we instead use the bounding box of $\mathcal{P}$ for the normalization, we observe a less detailed reconstruction of the well-observed foreground.
We further show that using smaller hashmaps for the multi-resolution hash grid has a modest impact on speed and quality but a major impact on model size.
Note that the impact on speed mostly depends on the amount of available L2 cache on the used GPU, which was extensively discussed by \citet{mueller2022instant}

\myparagraph{View-Dependence.}
We also analyze different representations for capturing view-dependent effects.
Not modeling view-dependence at all produces the worst results and using one spherical harmonics (SH) degree less impacts quality only on a few scenes.
Alternatives to SH such as spherical Gaussians (SG)~\cite{yu2021plenoctrees, yariv2023bakedsdf} and anisotropic spherical Gaussians (ASG)~\cite{xu2013anisotropic} perform worse in our tests.
For SG, we encode the lobe width inside the length of the mean vector, which results in $25$ and $32$ parameters per point with three and four lobes respectively.
For ASG, we use a 6D feature vector representing a rotation matrix~\cite{zhou2019continuity} for each lobe and use the basis vectors of this matrix to model the anisotropic extent.
This results in $15$ parameters per point for a configuration with a single ASG lobe.
We would like to emphasize that we find the ASG representation in particular much more elegant and compact as it uses more than $2\times$ less features per point and is theoretically less likely to overfit to the training views.
However, we observe that it is much harder to optimize using gradient descent compared to SH.

\myparagraph{U-Net.}
Lastly, we analyze the impact of using smaller U-Net configurations \wrt speed and image quality.
Using less initial filters considerably speeds up training and rendering at the cost of image quality.
In contrast, using only two layers barely has any impact on speed but reduces image quality significantly.

\myparagraph{Other.}
We show effect of using even less samples (4M) and observe much faster training and rendering with only slightly reduced image quality.
Here, we also want to highlight that for ablation J, which uses no post-processing but $16$M samples instead, the rendering speed is lower than with 8M samples and post-processing enabled.
Reducing the number of samples to $8$M results in $2.51$ hours training time and $10.9$ fps.
In combination, these two observations suggest finding a method for hole-filling that works with fewer samples without requiring a CNN could lead to a remarkably powerful point-based representation in the future.

\section{Per-Scene Results}
\label{sec:per_scene_results}
\myparagraph{Mip-NeRF360.~\cite{barron2022mipnerf360}}
We show per-scene image quality metrics and visual comparisons in \cref{tab:per_scene_metrics_m360} and \cref{fig:vis_comp_m360} respectively.
We follow the common practice of using $4\times$ downsampling for the five outdoor scenes and $2\times$ downsampling for the four indoor scenes.
Consequently, reconstructions and evaluation use images with roughly one megapixel (MP) resolution for outdoor and $1.5$ MP resolution for indoor scenes.
Importantly, we directly use the downsampled images provided with the dataset as we observe that these are of considerably higher quality compared to the output of commonly used downsampling implementations, e.g., those provided by OpenCV, PIL, and PyTorch$\slash$Torchvision.
In our tests, the used downsampling algorithm had a non-negligible influence on results.
When using the downsampled images \citet{barron2022mipnerf360} obtained using ImageMagick, we notice more accurate recovery of fine details but up to $0.5$ db lower PSNR values.
Intuitively this makes sense: A better downsampling algorithm leads to more details that \textit{can} be recovered but also more details that \textit{must} be recovered to achieve good metrics.
However, we suggest that a detailed study on the influence of using different downsampling algorithms in the context of novel view synthesis could be interesting.

\myparagraph{Tanks and Temples.~\cite{Knapitsch2017}}
We show per-scene image quality metrics and visual comparisons in \cref{tab:per_scene_metrics_tt} and \cref{fig:vis_comp_tt} respectively.
We evaluate on all eight scenes of the \textit{intermediate} set and use no downsampling, which results in images that have a resolution of roughly two MP.
As the Tanks and Temples dataset does not include camera parameters, we estimate them using COLMAP~\cite{schoenberger2016colmap1, schoenberger2016colmap2} and use the resulting calibration for all methods$\slash$experiments.

\myparagraph{About LPIPS and the VGG Loss.}
The most commonly used version of the Learned Perceptual Image Patch Similarity (LPIPS)~\cite{zhang2018lpips} metric uses a VGG-16~\cite{simonyan2014vgg} architecture, where $16$ quantifies the number of convolution layers in the employed neural network.
When inspecting our loss function in combination with our quantitative evaluation, it is easy to assume that our LPIPS results are so good specifically because we use a VGG-based loss term ($\mathcal{L}_\text{VGG}$).
We highlight that -- similar to prior works~\cite{vgg19loss,ruckert2022adop,franke2024trips} -- our VGG loss uses the VGG-19~\cite{simonyan2014vgg} architecture.
This is a key difference compared to both the original version of the loss by \citet{vgg16loss} and the used LPIPS metric.
Our ablation study (see \cref{tab:extra_ablations}) clearly shows that using the VGG loss improves results \wrt all three reported quality metrics.
To further validate that exploitation of similarities in the VGG-16 and VGG-19 architectures is not a deciding factor for our improved results, we include all three available versions of LPIPS (LPIPS\textsubscript{vgg}$\slash$LPIPS\textsubscript{alex}$\slash$LPIPS\textsubscript{squeeze}) in the per-scene result tables (\cref{tab:per_scene_metrics_m360} and \ref{tab:per_scene_metrics_tt}).
Our method achieves the best average performance for all versions of LPIPS on both datasets.

\section{Perceptual Experiment}\label{sec:experiment}
The objective of our perceptual experiment is to compare, from an observer's perspective, the novel view results of our method and Zip-NeRF~\cite{barron2023ICCV} (the method achieving the best quality metrics' scores among the compared-against methods).
This necessitates the presentation of a multitude of stimuli, the distinctions between which are frequently nuanced. It is also crucial to acknowledge that the quality of the results we seek to quantify cannot be accurately represented on a linear scale~\cite{Kendall1940}. This underscores the limitations of ranking methods in this context. Accordingly, the paired comparisons technique was selected for use in this experiment. This technique requires participants to view two images simultaneously and then select the one that more closely aligns with the task question. This resulted in a two-alternatives forced-choice (2AFC) preference task, in which each result was compared to the corresponding other method's results for the same image. 

\subsection{Experimental Design}

\myparagraph{Stimuli.}
Consistently with the analysis of the main paper, the stimuli were selected from the well established datasets Mip-NeRF360~\cite{barron2022mipnerf360} and Tanks and Temples~\cite{Knapitsch2017}. In order to maintain the number of trials of the experiment under a reasonable number that allows for a single participant to perform a complete test while maintaining the necessary level of attention, we selected a set of three novel views per scene, for all scenes in both considered datasets. The criteria for selection aimed to cover the maximum possible diversity in image attributes (i.e., background$\slash$foreground objects, recurring texture, scene composition, lighting, faces$\slash$people, lines$\slash$clear edges, view angle$\slash$focus depth, textual elements) for a given scene. For those scenes were the attributes were not sufficiently sampled, we extended the number of considered novel views (i.e., \textit{Room}: $4$, \textit{Lighthouse}: $4$, \textit{M60}: $5$). We then gathered the corresponding results of both Zip-NeRF~\cite{barron2023ICCV} and our method, for a total of $55$ paired comparisons (see~\cref{fig:exp-Stimuli}).

\myparagraph{Participants.}
A total of $17$ participants were included in the study, with an age range of $22$ to $55$ years and a gender distribution of $4$ females, $1$ non-available, and $12$ males. All participants reported normal or corrected-to-normal vision. Additionally, the majority of participants indicated an intermediate level of proficiency in computer graphics and image processing, as measured on a scale of $1$ to $4$, with $1$ indicating no experience and $4$ indicating advanced proficiency.

\myparagraph{Procedure.}
The participants performed the experiment in-situ one at a time. Before starting the experiment, each participant signed an informed consent form. They were then given an explanation on the experimental task and provided with the options of asking any questions. However, they were neither informed of the research question behind the experiment, nor of the nature (e.g., captured vs. generated) of the images they were presented with. 
The participant was asked to sit roughly $50$ cm in front of a 27” UHD monitor (at a resolution of $3840\times2160$) in a semi-dark room.
After the experimenter gathered the participant's demographic data, the partaker was presented with a screen describing once again the instructions. Before the experimenter left the room, the participants were given another chance to ask any further questions.
Each trial started by presenting two images side by side over a black background. Participants were not explicitly instructed to focus their attention on anything in particular, but to just select which of the two images they found more appealing. The participants were able to enter their answers by clicking on the desired image without having any restriction regarding the time to make their choice; once a response was entered, the next trial started. 
Both the order of the pairs of stimuli, and their position on the screen (left vs. right) was fully randomized, with each participant receiving a different random order. The experiment was controlled by Psychophysics Toolbox Version 3.0.15 (PTB-3) \cite{psytb1,psytb2}, and the mean time to complete it was of $21$ minutes.

\subsection{Analysis}

\begin{figure*}[th]
    \centering
    \vspace{-15pt}
    \includegraphics[height=.99\textheight, keepaspectratio]{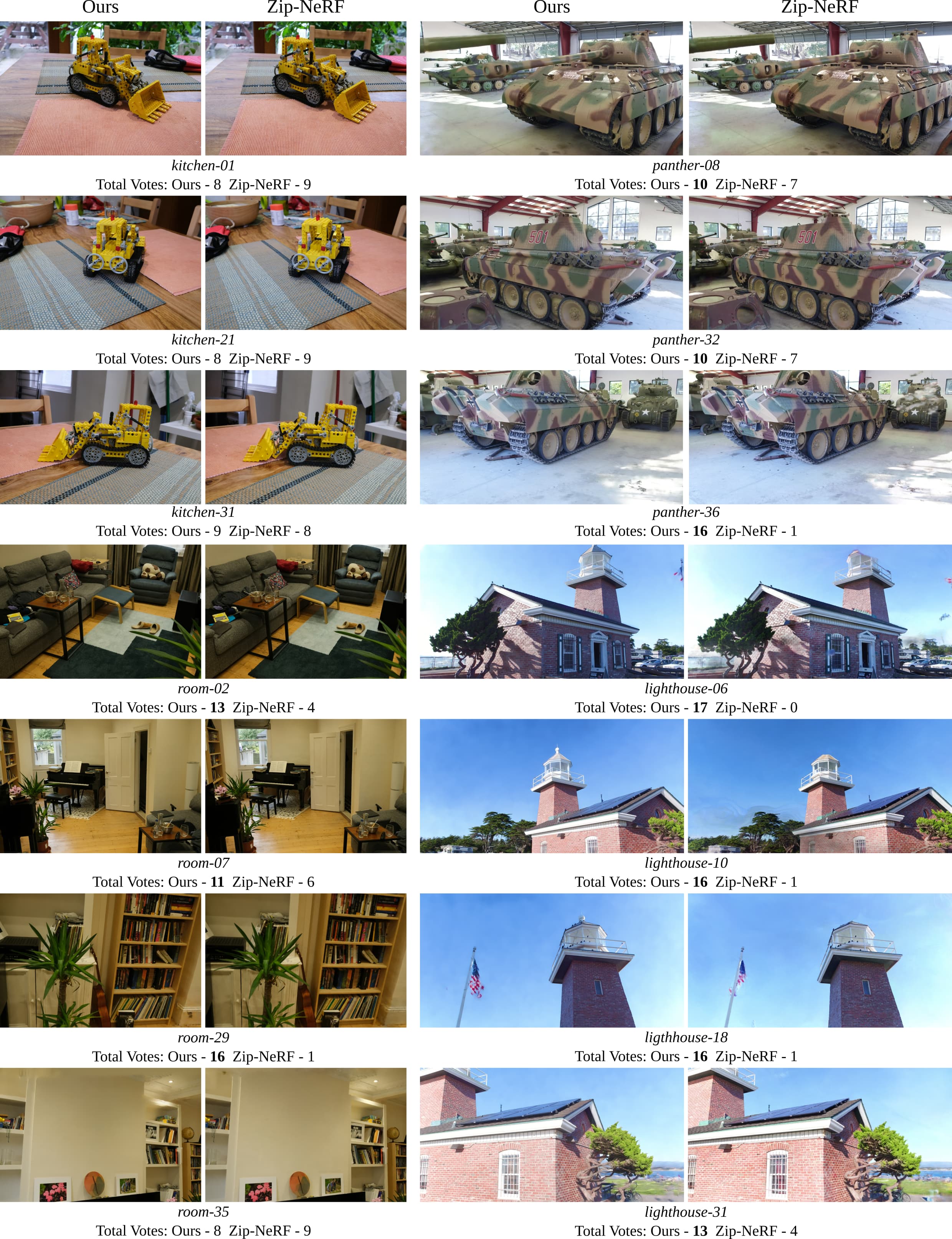}
    \vspace{-6pt}
\caption{
\textbf{Experimental Stimuli}. All images generated with Zip-NeRF~\cite{barron2023ICCV} and our method used as stimuli for our perceptual experiment. The two columns of the left belong to the results on the Mip-NeRF360 dataset~\cite{barron2022mipnerf360}, while the two on the right are from the Tanks and Temples dataset~\cite{Knapitsch2017}. Under each pair of stimuli, we report the scene name, frame number, and total votes gathered per method (max=17).
}\label{fig:exp-Stimuli}%
\end{figure*}

To assess not only the efficacy of each method, but also the consistency of participant responses, we employed the linked-paired comparison design~\cite{david1963method}. Accordingly, the methods are ranked according to the number of times they are preferred over the other method. The total number of votes a pair of stimuli received is displayed in~\cref{fig:exp-Stimuli}, while the votes over datasets and over all stimuli are displayed in~\cref{tab:Exp}. These results favorably rank our method over Zip-NeRF in all scenarios.
In order to ascertain the true meaning of this ranking, a significance test of the score differences is performed. In order to achieve this objective, it is necessary to identify a value, designated as $R'$, which represents the minimum variance-normalized range of scores within each group. This necessitates the computation of $R'$ such that $P[R \ge R'] \le \alpha$, where the confidence level, set at $0.01$, is represented by $\alpha$. Subsequently, in accordance with the methodology proposed by Herbert A. David~\cite{david1963method}, we can derive $R'$ from the following equation:
\begin{equation}
    P(W_{t,\alpha} \geq \frac{2R' - 0.5}{\sqrt{mt}}),
\end{equation}
where $m$ is the number of participants, $t$ is the number of methods to be compared, and $W_{t,\alpha}$ has been previously tabulated by Pearson and Hartley~\cite{Pearson1966Tables}. In our case, $W_{2,0.01} = 3.645$,  which leads to $R'_{Exp} = 10.87691$ for our experiment.

Since the score differences between all ranked groups exceed the aforementioned $R'$, we can conclude that they are all statistically significant. Thus, the ranking creates two distinguishable groups within our experimental context. 

\begin{table}[hbt]
    \centering
    \setlength\tabcolsep{10 pt}
    {\scriptsize
    \begin{tabular}{@{}llcc}
         \toprule
         Dataset & Method & \#Votes & Ranking\\
         \midrule
         \multirow{2}{*}{Mip-NeRF360~\cite{barron2022mipnerf360}}& Ours & 331 & $1^\text{st}$\\
         &Zip-NeRF~\cite{barron2023ICCV} & 145 & $2^\text{nd}$\\
         \midrule
         \multirow{2}{*}{Tanks \& Temples~\cite{Knapitsch2017}} &Ours & 318 &  $1^\text{st}$\\
         &Zip-NeRF~\cite{barron2023ICCV} & 141 & $2^\text{nd}$\\
         \midrule
         \multirow{2}{*}{All Stimuli} & Ours & 649 &  $1^\text{st}$\\
         &Zip-NeRF~\cite{barron2023ICCV} & 286 & $2^\text{nd}$\\
         \bottomrule
        \end{tabular}
    }
    \caption{Perceptual ranking of the compared methods in our experiment with $17$ participants. All rankings are statistically significant.
    }\label{tab:Exp}
\end{table}

\begin{table*}[ht]
\centering
\vspace{1cm}
\setlength\tabcolsep{11.4pt}
{\scriptsize
\begin{tabular}{@{}lccccc|cccc|c}
\toprule
\multicolumn{11}{c}{LPIPS\textsubscript{vgg}$\downarrow$ on Mip-NeRF360~\cite{barron2022mipnerf360}\vspace{2pt}}\\
Method & \textit{Bicycle} & \textit{Flowers} & \textit{Garden} & \textit{Stump} & \textit{Treehill} & \textit{Bonsai} & \textit{Counter} & \textit{Kitchen} & \textit{Room} & \textit{Average} \\
\midrule
Instant-NGP~\cite{mueller2022instant} & 0.478 & 0.466 & 0.289 & 0.474 & 0.496 & 0.258 & 0.368 & 0.249 & 0.340 & 0.380 \\
ADOP~\cite{ruckert2022adop}           & 0.250 & 0.361 & 0.203 & 0.305 & 0.354 & 0.223 & 0.264 & 0.221 & 0.241 & 0.259 \\
TRIPS~\cite{franke2024trips}          & \cellcolor{zwei}0.223& \cellcolor{drei}0.318& 0.183 & 0.309 & \cellcolor{drei}0.308& \cellcolor{zwei}0.153& \cellcolor{zwei}0.206& \cellcolor{drei}0.154& \cellcolor{zwei}0.197& \cellcolor{zwei}0.213\\
3DGS~\cite{kerbl3Dgaussians}          & 0.229 & 0.366 & \cellcolor{zwei}0.118& \cellcolor{drei}0.244& 0.367 & 0.253 & 0.262 & 0.158 & 0.289 & 0.254 \\
Zip-NeRF~\cite{barron2023ICCV}        & \cellcolor{drei}0.228& \cellcolor{zwei}0.309& \cellcolor{drei}0.127& \cellcolor{zwei}0.236& \cellcolor{zwei}0.281& \cellcolor{drei}0.196& \cellcolor{drei}0.223& \cellcolor{zwei}0.134& \cellcolor{drei}0.238& \cellcolor{drei}0.219\\
Ours                                  & \cellcolor{eins}0.161& \cellcolor{eins}0.212& \cellcolor{eins}0.086& \cellcolor{eins}0.173& \cellcolor{eins}0.215& \cellcolor{eins}0.137& \cellcolor{eins}0.184& \cellcolor{eins}0.123& \cellcolor{eins}0.187& \cellcolor{eins}0.164\\
\midrule
Ours (16M)                            & 0.171 & 0.220 & 0.090 & 0.190 & 0.226 & 0.149 & 0.187 & 0.126 & 0.195 & 0.173 \\
Ours (8M)                             & 0.183 & 0.246 & 0.098 & 0.210 & 0.242 & 0.163 & 0.212 & 0.134 & 0.202 & 0.188 \\
Ours (pre-ex.)                        & 0.220 & 0.254 & 0.113 & 0.278 & 0.276 & 0.156 & 0.224 & 0.137 & 0.200 & 0.207 \\
\bottomrule
\end{tabular}
\begin{tabular}{@{}lccccc|cccc|c}
\toprule
\multicolumn{11}{c}{SSIM$\uparrow$ on Mip-NeRF360~\cite{barron2022mipnerf360}\vspace{2pt}}\\
Method & \textit{Bicycle} & \textit{Flowers} & \textit{Garden} & \textit{Stump} & \textit{Treehill} & \textit{Bonsai} & \textit{Counter} & \textit{Kitchen} & \textit{Room} & \textit{Average} \\
\midrule
Instant-NGP~\cite{mueller2022instant} & 0.513 & 0.485 & 0.706 & 0.591 & 0.544 & 0.904 & 0.816 & 0.856 & 0.870 & 0.698 \\
ADOP~\cite{ruckert2022adop}           & 0.665 & 0.494 & 0.741 & 0.666 & 0.556 & 0.818 & 0.769 & 0.737 & 0.839 & 0.723 \\
TRIPS~\cite{franke2024trips}          & 0.704 & 0.502 & 0.773 & 0.681 & 0.591 & 0.899 & 0.845 & 0.850 & 0.883 & 0.778 \\
3DGS~\cite{kerbl3Dgaussians}          & \cellcolor{drei}0.770& \cellcolor{drei}0.602& \cellcolor{zwei}0.869& \cellcolor{drei}0.774& \cellcolor{drei}0.637& \cellcolor{drei}0.938& \cellcolor{zwei}0.905& \cellcolor{drei}0.921& \cellcolor{drei}0.913& \cellcolor{drei}0.814\\
Zip-NeRF~\cite{barron2023ICCV}        & \cellcolor{zwei}0.772& \cellcolor{zwei}0.637& \cellcolor{drei}0.863& \cellcolor{zwei}0.788& \cellcolor{zwei}0.674& \cellcolor{zwei}0.952& \cellcolor{zwei}0.905& \cellcolor{zwei}0.929& \cellcolor{zwei}0.929& \cellcolor{zwei}0.828\\
Ours                                  & \cellcolor{eins}0.805& \cellcolor{eins}0.668& \cellcolor{eins}0.886& \cellcolor{eins}0.826& \cellcolor{eins}0.702& \cellcolor{eins}0.954& \cellcolor{eins}0.912& \cellcolor{eins}0.932& \cellcolor{eins}0.932& \cellcolor{eins}0.847\\
\midrule
Ours (16M)                            & 0.796 & 0.655 & 0.884 & 0.811 & 0.695 & 0.950 & 0.912 & 0.933 & 0.930 & 0.841 \\
Ours (8M)                             & 0.783 & 0.645 & 0.876 & 0.790 & 0.688 & 0.942 & 0.896 & 0.927 & 0.926 & 0.830 \\
Ours (pre-ex.)                        & 0.731 & 0.617 & 0.852 & 0.711 & 0.645 & 0.941 & 0.880 & 0.923 & 0.920 & 0.802 \\
\bottomrule
\end{tabular}
\begin{tabular}{@{}lccccc|cccc|c}
\toprule
\multicolumn{11}{c}{PSNR$\uparrow$ on Mip-NeRF360~\cite{barron2022mipnerf360}\vspace{2pt}}\\
Method & \textit{Bicycle} & \textit{Flowers} & \textit{Garden} & \textit{Stump} & \textit{Treehill} & \textit{Bonsai} & \textit{Counter} & \textit{Kitchen} & \textit{Room} & \textit{Average} \\
\midrule
Instant-NGP~\cite{mueller2022instant} & 22.21 & 20.68 & 25.14 & 23.47 & 22.42 & 30.69 & 26.69 & 29.48 & 29.71 & 25.61 \\
ADOP~\cite{ruckert2022adop}           & 22.60 & 19.68 & 24.85 & 24.18 & 20.99 & 24.33 & 23.09 & 23.61 & 25.97 & 23.54 \\
TRIPS~\cite{franke2024trips}          & 23.47 & 19.44 & 25.38 & 24.17 & 22.04 & 28.71 & 27.00 & 27.66 & 29.07 & 25.94 \\
3DGS~\cite{kerbl3Dgaussians}          & \cellcolor{drei}25.25& \cellcolor{drei}21.52& \cellcolor{drei}27.41& \cellcolor{drei}26.55& \cellcolor{drei}22.49& \cellcolor{drei}31.98& \cellcolor{drei}28.69& \cellcolor{drei}30.32& \cellcolor{drei}30.63& \cellcolor{drei}27.20\\ 
Zip-NeRF~\cite{barron2023ICCV}        & \cellcolor{zwei}25.85& \cellcolor{zwei}22.33& \cellcolor{zwei}28.22& \cellcolor{zwei}27.35& \cellcolor{zwei}23.95& \cellcolor{eins}34.79& \cellcolor{zwei}29.12& \cellcolor{eins}32.36& \cellcolor{eins}33.04& \cellcolor{eins}28.56\\
Ours                                  & \cellcolor{eins}26.26& \cellcolor{eins}22.41& \cellcolor{eins}28.52& \cellcolor{eins}27.71& \cellcolor{eins}24.16& \cellcolor{zwei}33.89& \cellcolor{eins}29.38 & \cellcolor{zwei}31.81& \cellcolor{zwei}32.89& \cellcolor{eins}28.56\\
\midrule
Ours (16M)                            & 26.09 & 22.19 & 28.34 & 27.44 & 24.20 & 33.29 & 29.20 & 31.61 & 32.91 & 28.36 \\
Ours (8M)                             & 25.77 & 22.12 & 28.09 & 26.85 & 24.05 & 31.90 & 28.39 & 30.92 & 32.36 & 27.83 \\
Ours (pre-ex.)                        & 24.45 & 21.60 & 26.78 & 25.35 & 23.41 & 31.72 & 26.69 & 30.47 & 31.22 & 26.85 \\
\bottomrule
\end{tabular}
\begin{tabular}{@{}lccccc|cccc|c}
\toprule
\multicolumn{11}{c}{LPIPS\textsubscript{alex}$\downarrow$ on Mip-NeRF360~\cite{barron2022mipnerf360}\vspace{2pt}}\\
Method & \textit{Bicycle} & \textit{Flowers} & \textit{Garden} & \textit{Stump} & \textit{Treehill} & \textit{Bonsai} & \textit{Counter} & \textit{Kitchen} & \textit{Room} & \textit{Average} \\
\midrule
Instant-NGP~\cite{mueller2022instant} & 0.389 & 0.385 & 0.221 & 0.342 & 0.468 & 0.123 & 0.244 & 0.137 & 0.206 & 0.279 \\
ADOP~\cite{ruckert2022adop}           & 0.166 & 0.280 & 0.132 & 0.208 & 0.278 & 0.129 & 0.177 & 0.151 & 0.164 & 0.177 \\
TRIPS~\cite{franke2024trips}          & \cellcolor{zwei}0.138& \cellcolor{drei}0.218& 0.103 & 0.205 & \cellcolor{zwei}0.218& \cellcolor{zwei}0.073& \cellcolor{zwei}0.124& 0.102 & \cellcolor{zwei}0.123& \cellcolor{zwei}0.133\\
3DGS~\cite{kerbl3Dgaussians}          & 0.165 & 0.340 & \cellcolor{zwei}0.074& \cellcolor{drei}0.151& 0.318 & 0.132 & 0.161 & \cellcolor{drei}0.099& 0.170 & 0.179 \\
Zip-NeRF~\cite{barron2023ICCV}        & \cellcolor{drei}0.161& \cellcolor{zwei}0.213& \cellcolor{drei}0.085& \cellcolor{zwei}0.129& \cellcolor{drei}0.223& \cellcolor{drei}0.092& \cellcolor{drei}0.129& \cellcolor{eins}0.082& \cellcolor{drei}0.133& \cellcolor{drei}0.139\\
Ours                                  & \cellcolor{eins}0.124& \cellcolor{eins}0.169& \cellcolor{eins}0.057& \cellcolor{eins}0.112& \cellcolor{eins}0.163& \cellcolor{eins}0.070& \cellcolor{eins}0.123& \cellcolor{zwei}0.083& \cellcolor{eins}0.116& \cellcolor{eins}0.113\\
\midrule
Ours (16M)                            & 0.137 & 0.183 & 0.060 & 0.131 & 0.173 & 0.080 & 0.125 & 0.087 & 0.120 & 0.122 \\
Ours (8M)                             & 0.152 & 0.213 & 0.067 & 0.161 & 0.214 & 0.098 & 0.153 & 0.096 & 0.129 & 0.143 \\
Ours (pre-ex.)                        & 0.170 & 0.207 & 0.078 & 0.192 & 0.232 & 0.090 & 0.156 & 0.095 & 0.128 & 0.150 \\
\bottomrule
\end{tabular}
\begin{tabular}{@{}lccccc|cccc|c}
\toprule
\multicolumn{11}{c}{LPIPS\textsubscript{squeeze}$\downarrow$ on Mip-NeRF360~\cite{barron2022mipnerf360}\vspace{2pt}}\\
Method & \textit{Bicycle} & \textit{Flowers} & \textit{Garden} & \textit{Stump} & \textit{Treehill} & \textit{Bonsai} & \textit{Counter} & \textit{Kitchen} & \textit{Room} & \textit{Average} \\
\midrule
Instant-NGP~\cite{mueller2022instant} & 0.260 & 0.269 & 0.137 & 0.254 & 0.285 & 0.105 & 0.185 & 0.112 & 0.155 & 0.196 \\
ADOP~\cite{ruckert2022adop}           & 0.104 & 0.179 & 0.091 & 0.139 & 0.155 & \cellcolor{drei}0.084& 0.111 & 0.096 & \cellcolor{drei}0.103& 0.112 \\
TRIPS~\cite{franke2024trips}          & \cellcolor{zwei}0.089& \cellcolor{zwei}0.131& 0.072 & 0.141 & \cellcolor{drei}0.123& \cellcolor{eins}0.049& \cellcolor{eins}0.080& \cellcolor{drei}0.065& \cellcolor{zwei}0.078& \cellcolor{zwei}0.084\\
3DGS~\cite{kerbl3Dgaussians}          & 0.108 & 0.246 & \cellcolor{zwei}0.051& \cellcolor{drei}0.112& 0.213 & 0.127 & 0.134 & 0.078 & 0.148 & 0.135 \\
Zip-NeRF~\cite{barron2023ICCV}        & \cellcolor{drei}0.095& \cellcolor{drei}0.136& \cellcolor{drei}0.053& \cellcolor{zwei}0.102& \cellcolor{zwei}0.117& 0.085 & \cellcolor{drei}0.093& \cellcolor{zwei}0.061& \cellcolor{drei}0.103& \cellcolor{drei}0.094\\
Ours                                  & \cellcolor{eins}0.073& \cellcolor{eins}0.094& \cellcolor{eins}0.037& \cellcolor{eins}0.079& \cellcolor{eins}0.090& \cellcolor{zwei}0.050& \cellcolor{zwei}0.082& \cellcolor{eins}0.057& \cellcolor{eins}0.076& \cellcolor{eins}0.071\\
\midrule
Ours (16M)                            & 0.081 & 0.104 & 0.040 & 0.093 & 0.096 & 0.058 & 0.084 & 0.059 & 0.079 & 0.077 \\
Ours (8M)                             & 0.091 & 0.122 & 0.045 & 0.111 & 0.113 & 0.069 & 0.104 & 0.064 & 0.086 & 0.089 \\
Ours (pre-ex.)                        & 0.102 & 0.119 & 0.049 & 0.133 & 0.120 & 0.061 & 0.103 & 0.063 & 0.082 & 0.092 \\
\bottomrule
\end{tabular}
}
\caption{
\textbf{Per-scene image quality metrics} for the Mip-NeRF360 dataset~\cite{barron2022mipnerf360} separated into outdoor and indoor scenes.
The three best results are highlighted in \textcolor{einsText}{\textbf{green}} in descending order of saturation.
}\label{tab:per_scene_metrics_m360}%
\vspace{1cm}
\end{table*}

\begin{figure*}[ht]
\vspace{.1cm}
    \centering
    \setlength\tabcolsep{0pt}
    \begin{tabular}{*{5}{p{0.2\linewidth}<{\centering}}}
    \scriptsize TRIPS & \scriptsize 3DGS & \scriptsize Zip-NeRF & \scriptsize Ours & \scriptsize Ground Truth
    \end{tabular}
    \includegraphics[width=\linewidth, keepaspectratio]{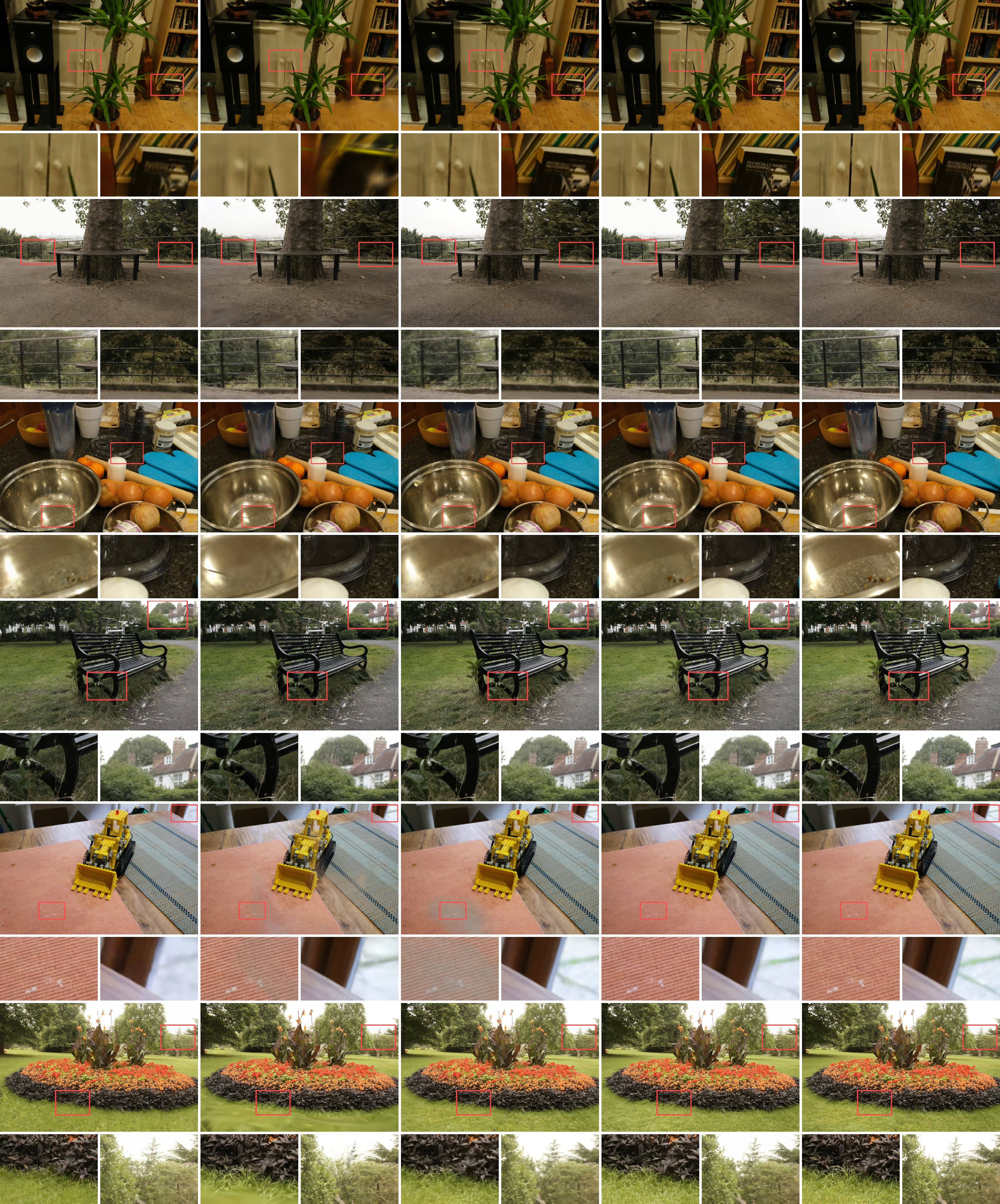}
\caption{%
\textbf{Visual comparisons} for multiple scenes of the Mip-NeRF360 dataset~\cite{barron2022mipnerf360}.
}\label{fig:vis_comp_m360}
\end{figure*}

\begin{table*}[ht]
\vspace{1cm}
\centering
\setlength\tabcolsep{13.1pt}
{\scriptsize
\begin{tabular}{@{}lcccccccc|c}
\toprule
\multicolumn{10}{c}{LPIPS\textsubscript{vgg}$\downarrow$ on Tanks and Temples~\cite{Knapitsch2017}\vspace{2pt}}\\
Method & \textit{Family} & \textit{Francis} & \textit{Horse} & \textit{Lighthouse} & \textit{M60} & \textit{Panther} & \textit{Playground} & \textit{Train} & \textit{Average} \\
\midrule
Instant-NGP~\cite{mueller2022instant} & 0.413 & 0.439 & 0.458 & 0.439 & 0.367 & 0.355 & 0.547 & 0.487 & 0.438 \\
ADOP~\cite{ruckert2022adop}           & 0.203 & \cellcolor{zwei}0.233& 0.201 & \cellcolor{zwei}0.242& 0.225 & 0.219 & \cellcolor{drei}0.231& 0.302 & 0.236 \\
TRIPS~\cite{franke2024trips}          & \cellcolor{drei}0.176& \cellcolor{drei}0.266& \cellcolor{drei}0.182& \cellcolor{drei}0.277& \cellcolor{zwei}0.204& \cellcolor{zwei}0.191& \cellcolor{zwei}0.222& \cellcolor{eins}0.267& \cellcolor{zwei}0.229\\
3DGS~\cite{kerbl3Dgaussians}          & 0.236 & 0.344 & 0.239 & 0.291 & 0.244 & 0.241 & 0.291 & 0.320 & 0.276 \\
Zip-NeRF~\cite{barron2023ICCV}        & \cellcolor{zwei}0.172& 0.270 & \cellcolor{zwei}0.181& 0.281 & \cellcolor{drei}0.212& \cellcolor{drei}0.217& 0.251 & \cellcolor{zwei}0.279& \cellcolor{drei}0.233\\
Ours                                  & \cellcolor{eins}0.115& \cellcolor{eins}0.227& \cellcolor{eins}0.134& \cellcolor{eins}0.217& \cellcolor{eins}0.182& \cellcolor{eins}0.178& \cellcolor{eins}0.172& \cellcolor{drei}0.287& \cellcolor{eins}0.189\\
\midrule
Ours (16M)                            & 0.125 & 0.241 & 0.146 & 0.236 & 0.239 & 0.220 & 0.195 & 0.356 & 0.220 \\
Ours (8M)                             & 0.141 & 0.253 & 0.159 & 0.275 & 0.304 & 0.267 & 0.215 & 0.398 & 0.252 \\
Ours (pre-ex.)                        & 0.174 & 0.292 & 0.207 & 0.265 & 0.263 & 0.232 & 0.237 & 0.421 & 0.261 \\
\bottomrule
\end{tabular}
\begin{tabular}{@{}lcccccccc|c}
\toprule
\multicolumn{10}{c}{SSIM$\uparrow$ on Tanks and Temples~\cite{Knapitsch2017}\vspace{2pt}}\\
Method & \textit{Family} & \textit{Francis} & \textit{Horse} & \textit{Lighthouse} & \textit{M60} & \textit{Panther} & \textit{Playground} & \textit{Train} & \textit{Average} \\
\midrule
Instant-NGP~\cite{mueller2022instant} & 0.729 & 0.812 & 0.733 & 0.759 & 0.810 & 0.840 & 0.550 & 0.666 & 0.737 \\
ADOP~\cite{ruckert2022adop}           & 0.807 & 0.860 & 0.842 & 0.782 & 0.835 & 0.859 & 0.785 & 0.667 & 0.802 \\
TRIPS~\cite{franke2024trips}          & 0.849 & 0.879 & 0.871 & 0.792 & 0.862 & 0.884 & 0.771 & 0.768 & 0.831 \\
3DGS~\cite{kerbl3Dgaussians}          & \cellcolor{drei}0.871& \cellcolor{drei}0.901& \cellcolor{drei}0.889& \cellcolor{zwei}0.834& \cellcolor{drei}0.901& \cellcolor{zwei}0.910& \cellcolor{drei}0.834& \cellcolor{zwei}0.791& \cellcolor{zwei}0.866\\
Zip-NeRF~\cite{barron2023ICCV}        & \cellcolor{zwei}0.893& \cellcolor{eins}0.918& \cellcolor{zwei}0.909& \cellcolor{eins}0.835& \cellcolor{eins}0.905& \cellcolor{drei}0.908& \cellcolor{zwei}0.846& \cellcolor{eins}0.813& \cellcolor{eins}0.878\\
Ours                                  & \cellcolor{eins}0.905& \cellcolor{zwei}0.915& \cellcolor{eins}0.914& \cellcolor{drei}0.833& \cellcolor{zwei}0.903& \cellcolor{eins}0.912& \cellcolor{eins}0.878& \cellcolor{drei}0.769& \cellcolor{eins}0.878\\
\midrule
Ours (16M)                            & 0.902 & 0.906 & 0.909 & 0.822 & 0.884 & 0.896 & 0.860 & 0.718 & 0.862 \\
Ours (8M)                             & 0.895 & 0.906 & 0.905 & 0.804 & 0.848 & 0.870 & 0.852 & 0.687 & 0.846 \\
Ours (pre-ex.)                        & 0.864 & 0.879 & 0.877 & 0.808 & 0.864 & 0.884 & 0.826 & 0.664 & 0.833 \\
\bottomrule
\end{tabular}
\begin{tabular}{@{}lcccccccc|c}
\toprule
\multicolumn{10}{c}{PSNR$\uparrow$ on Tanks and Temples~\cite{Knapitsch2017}\vspace{2pt}}\\
Method & \textit{Family} & \textit{Francis} & \textit{Horse} & \textit{Lighthouse} & \textit{M60} & \textit{Panther} & \textit{Playground} & \textit{Train} & \textit{Average} \\
\midrule
Instant-NGP~\cite{mueller2022instant} & 21.47 & 23.96 & 18.45 & 21.17 & 24.87 & 26.45 & 18.52 & \cellcolor{drei}19.72& 21.82 \\
ADOP~\cite{ruckert2022adop}           & 24.29 & 21.92 & 23.87 & 18.28 & 23.66 & 25.47 & 22.58 & 15.66 & 21.69 \\
TRIPS~\cite{franke2024trips}          & 24.03 & 20.06 & 23.45 & 18.09 & 25.52 & 27.73 & 24.10 & 18.79 & 22.62 \\
3DGS~\cite{kerbl3Dgaussians}          & \cellcolor{drei}25.05& \cellcolor{zwei}27.64& \cellcolor{drei}24.18& \cellcolor{drei}21.76& \cellcolor{zwei}27.82& \cellcolor{drei}28.35& \cellcolor{drei}25.65& \cellcolor{zwei}21.69& \cellcolor{drei}25.27\\
Zip-NeRF~\cite{barron2023ICCV}        & \cellcolor{zwei}28.05& \cellcolor{eins}29.55& \cellcolor{zwei}27.67& \cellcolor{eins}22.31& \cellcolor{eins}28.86& \cellcolor{eins}28.84& \cellcolor{zwei}26.62& \cellcolor{eins}22.10& \cellcolor{eins}26.75\\
Ours                                  & \cellcolor{eins}28.52& \cellcolor{drei}26.85& \cellcolor{eins}27.73& \cellcolor{zwei}21.83& \cellcolor{drei}27.77& \cellcolor{zwei}28.82& \cellcolor{eins}26.81& 19.12 & \cellcolor{zwei}25.93\\
\midrule
Ours (16M)                            & 28.05 & 26.61 & 27.10 & 21.40 & 26.55 & 27.55 & 25.61 & 18.11 & 25.12 \\
Ours (8M)                             & 27.92 & 26.09 & 26.86 & 21.08 & 24.56 & 25.99 & 26.04 & 17.81 & 24.54 \\
Ours (pre-ex.)                        & 26.20 & 24.23 & 24.53 & 20.96 & 25.22 & 26.43 & 25.15 & 16.98 & 23.71 \\
\bottomrule
\end{tabular}
\begin{tabular}{@{}lcccccccc|c}
\toprule
\multicolumn{10}{c}{LPIPS\textsubscript{alex}$\downarrow$ on Tanks and Temples~\cite{Knapitsch2017}\vspace{2pt}}\\
Method & \textit{Family} & \textit{Francis} & \textit{Horse} & \textit{Lighthouse} & \textit{M60} & \textit{Panther} & \textit{Playground} & \textit{Train} & \textit{Average} \\
\midrule
Instant-NGP~\cite{mueller2022instant} & 0.337 & 0.295 & 0.387 & 0.313 & 0.259 & 0.245 & 0.475 & 0.393 & 0.338 \\
ADOP~\cite{ruckert2022adop}           & 0.142 & \cellcolor{zwei}0.122& 0.138 & \cellcolor{drei}0.180& 0.148 & 0.128 & \cellcolor{drei}0.157& \cellcolor{drei}0.236& 0.158 \\
TRIPS~\cite{franke2024trips}          & \cellcolor{drei}0.114& 0.138 & \cellcolor{drei}0.115& 0.187 & \cellcolor{eins}0.124& \cellcolor{eins}0.101& \cellcolor{eins}0.133& \cellcolor{zwei}0.181& \cellcolor{drei}0.140\\
3DGS~\cite{kerbl3Dgaussians}          & 0.152 & 0.187 & 0.144 & 0.197 & 0.165 & 0.146 & 0.231 & 0.237 & 0.182 \\
Zip-NeRF~\cite{barron2023ICCV}        & \cellcolor{zwei}0.105& \cellcolor{drei}0.125& \cellcolor{zwei}0.099& \cellcolor{zwei}0.174& \cellcolor{zwei}0.125& \cellcolor{drei}0.118& 0.175 & \cellcolor{eins}0.171& \cellcolor{zwei}0.136\\
Ours                                  & \cellcolor{eins}0.073& \cellcolor{eins}0.113& \cellcolor{eins}0.072& \cellcolor{eins}0.152& \cellcolor{drei}0.136& \cellcolor{zwei}0.114& \cellcolor{zwei}0.145& 0.241 & \cellcolor{eins}0.131\\
\midrule
Ours (16M)                            & 0.087 & 0.126 & 0.085 & 0.176 & 0.192 & 0.160 & 0.174 & 0.323 & 0.165 \\
Ours (8M)                             & 0.105 & 0.139 & 0.099 & 0.222 & 0.266 & 0.213 & 0.196 & 0.371 & 0.201 \\
Ours (pre-ex.)                        & 0.128 & 0.182 & 0.148 & 0.200 & 0.205 & 0.151 & 0.200 & 0.385 & 0.200 \\
\bottomrule
\end{tabular}
\begin{tabular}{@{}lcccccccc|c}
\toprule
\multicolumn{10}{c}{LPIPS\textsubscript{squeeze}$\downarrow$ on Tanks and Temples~\cite{Knapitsch2017}\vspace{2pt}}\\
Method & \textit{Family} & \textit{Francis} & \textit{Horse} & \textit{Lighthouse} & \textit{M60} & \textit{Panther} & \textit{Playground} & \textit{Train} & \textit{Average} \\
\midrule
Instant-NGP~\cite{mueller2022instant} & 0.251 & 0.224 & 0.302 & 0.245 & 0.207 & 0.194 & 0.351 & 0.310 & 0.261 \\
ADOP~\cite{ruckert2022adop}           & 0.090 & \cellcolor{eins}0.085& 0.091 & \cellcolor{zwei}0.118& 0.100 & \cellcolor{drei}0.088& \cellcolor{drei}0.099& \cellcolor{drei}0.162& 0.106 \\
TRIPS~\cite{franke2024trips}          & \cellcolor{drei}0.076& 0.134 & \cellcolor{drei}0.080& 0.144 & \cellcolor{eins}0.086& \cellcolor{eins}0.071& \cellcolor{zwei}0.086& \cellcolor{zwei}0.136& \cellcolor{drei}0.105\\
3DGS~\cite{kerbl3Dgaussians}          & 0.120 & 0.180 & 0.122 & 0.164 & 0.137 & 0.127 & 0.175 & 0.196 & 0.153 \\
Zip-NeRF~\cite{barron2023ICCV}        & \cellcolor{zwei}0.074& \cellcolor{drei}0.104& \cellcolor{zwei}0.073& \cellcolor{drei}0.131& \cellcolor{drei}0.094& 0.091 & 0.117 & \cellcolor{eins}0.126& \cellcolor{zwei}0.101\\
Ours                                  & \cellcolor{eins}0.050& \cellcolor{zwei}0.093& \cellcolor{eins}0.052& \cellcolor{eins}0.112& \cellcolor{zwei}0.091& \cellcolor{zwei}0.080& \cellcolor{eins}0.083& 0.166 & \cellcolor{eins}0.091\\
\midrule
Ours (16M)                            & 0.058 & 0.102 & 0.060 & 0.126 & 0.133 & 0.113 & 0.101 & 0.222 & 0.114 \\
Ours (8M)                             & 0.071 & 0.111 & 0.071 & 0.159 & 0.183 & 0.148 & 0.116 & 0.263 & 0.140 \\
Ours (pre-ex.)                        & 0.083 & 0.132 & 0.102 & 0.139 & 0.141 & 0.110 & 0.114 & 0.273 & 0.137 \\
\bottomrule
\end{tabular}
}
\caption{%
\textbf{Per-scene image quality metrics} for the Tanks and Temples dataset~\cite{Knapitsch2017}.
The three best results are highlighted in \textcolor{einsText}{\textbf{green}} in descending order of saturation. 
}\label{tab:per_scene_metrics_tt}
\vspace{1cm}
\end{table*}

\begin{figure*}[ht]
\vspace{2cm}
    \centering
    \setlength\tabcolsep{0pt}
    \begin{tabular}{*{5}{p{0.2\linewidth}<{\centering}}}
    \scriptsize TRIPS & \scriptsize 3DGS & \scriptsize Zip-NeRF & \scriptsize Ours & \scriptsize Ground Truth
    \end{tabular}
    \includegraphics[width=\linewidth, keepaspectratio]{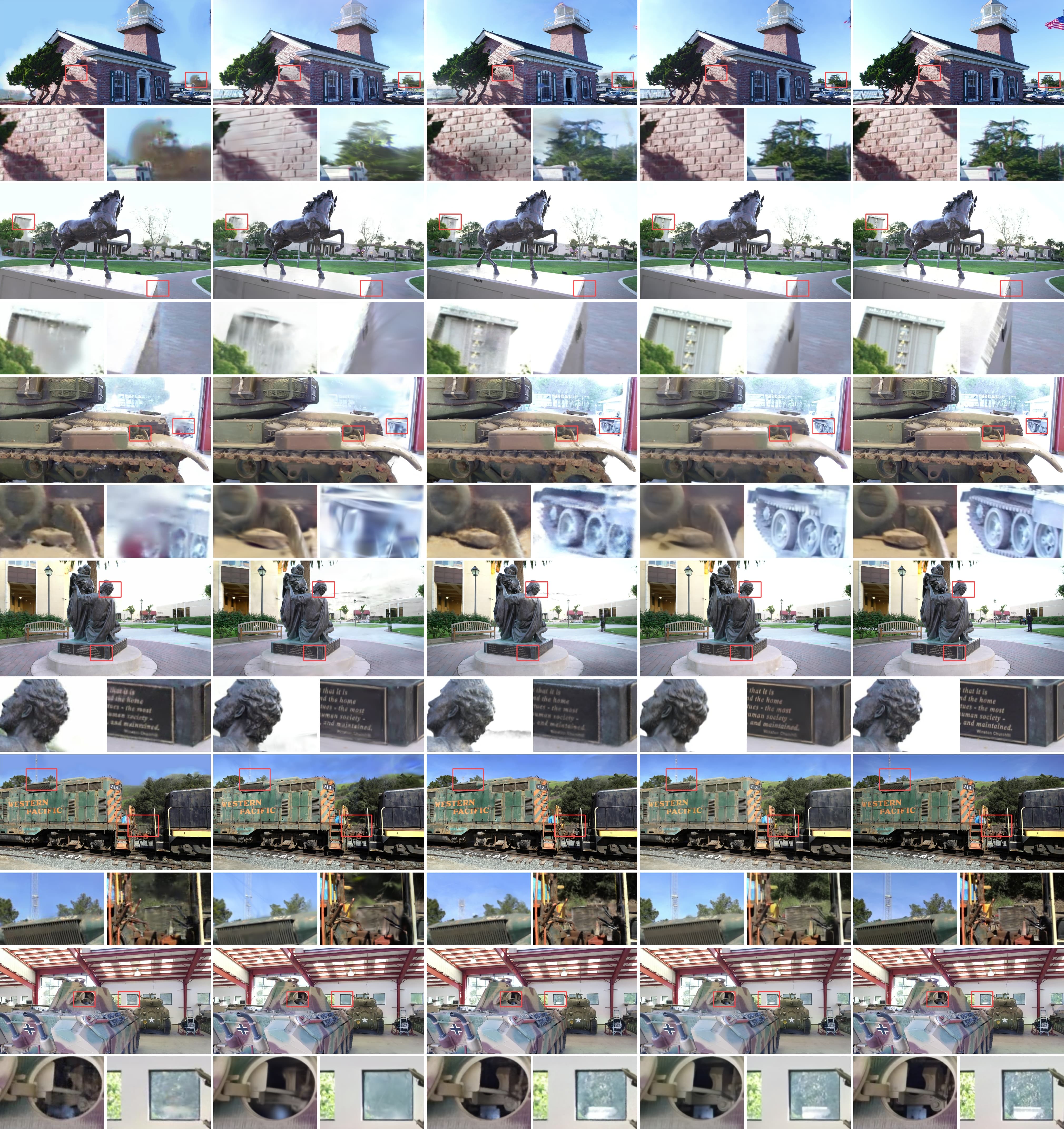}
\caption{
\textbf{Visual comparisons} for multiple scenes of the Tanks and Temples dataset~\cite{Knapitsch2017}.
}\label{fig:vis_comp_tt}%
\end{figure*}

\begin{figure*}[th]
    \centering
    \includegraphics[height=.99\textheight, keepaspectratio]{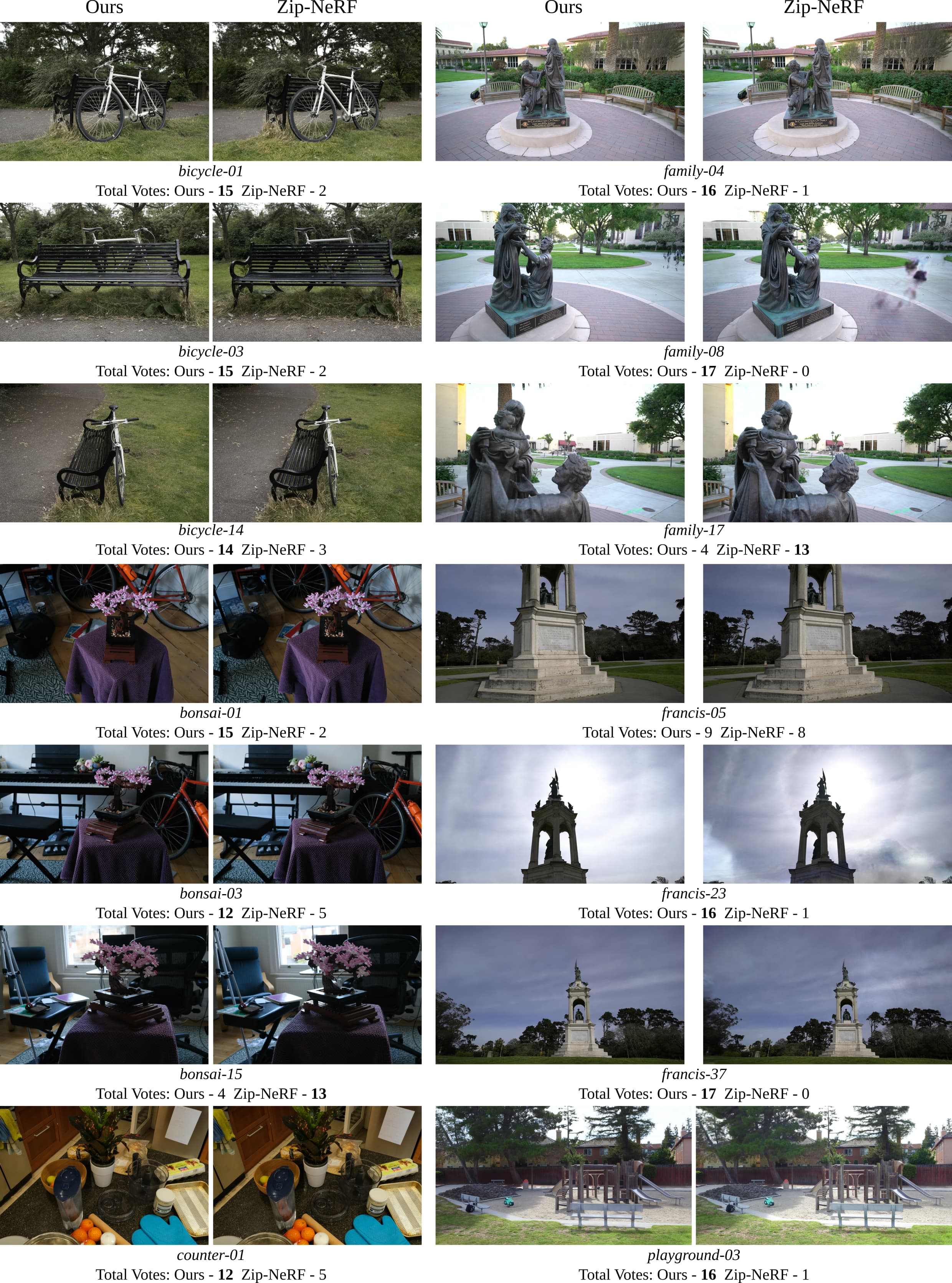}
\label{fig:exp1}%
\end{figure*}

\begin{figure*}[th]
    \centering
    \includegraphics[height=.99\textheight, keepaspectratio]{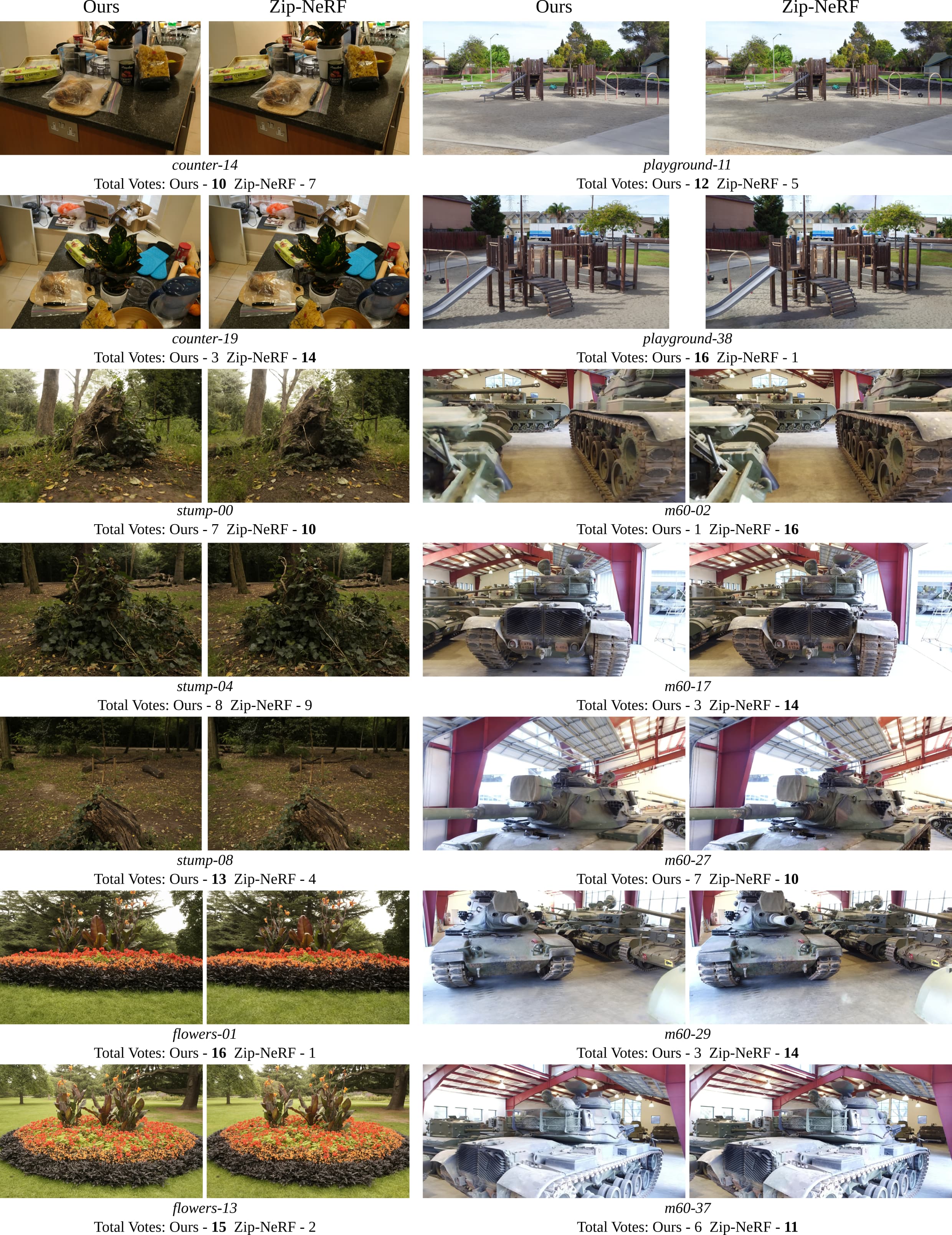}
\label{fig:exp2}%
\end{figure*}
\begin{figure*}[th]
    \centering
    \includegraphics[height=.99\textheight, keepaspectratio]{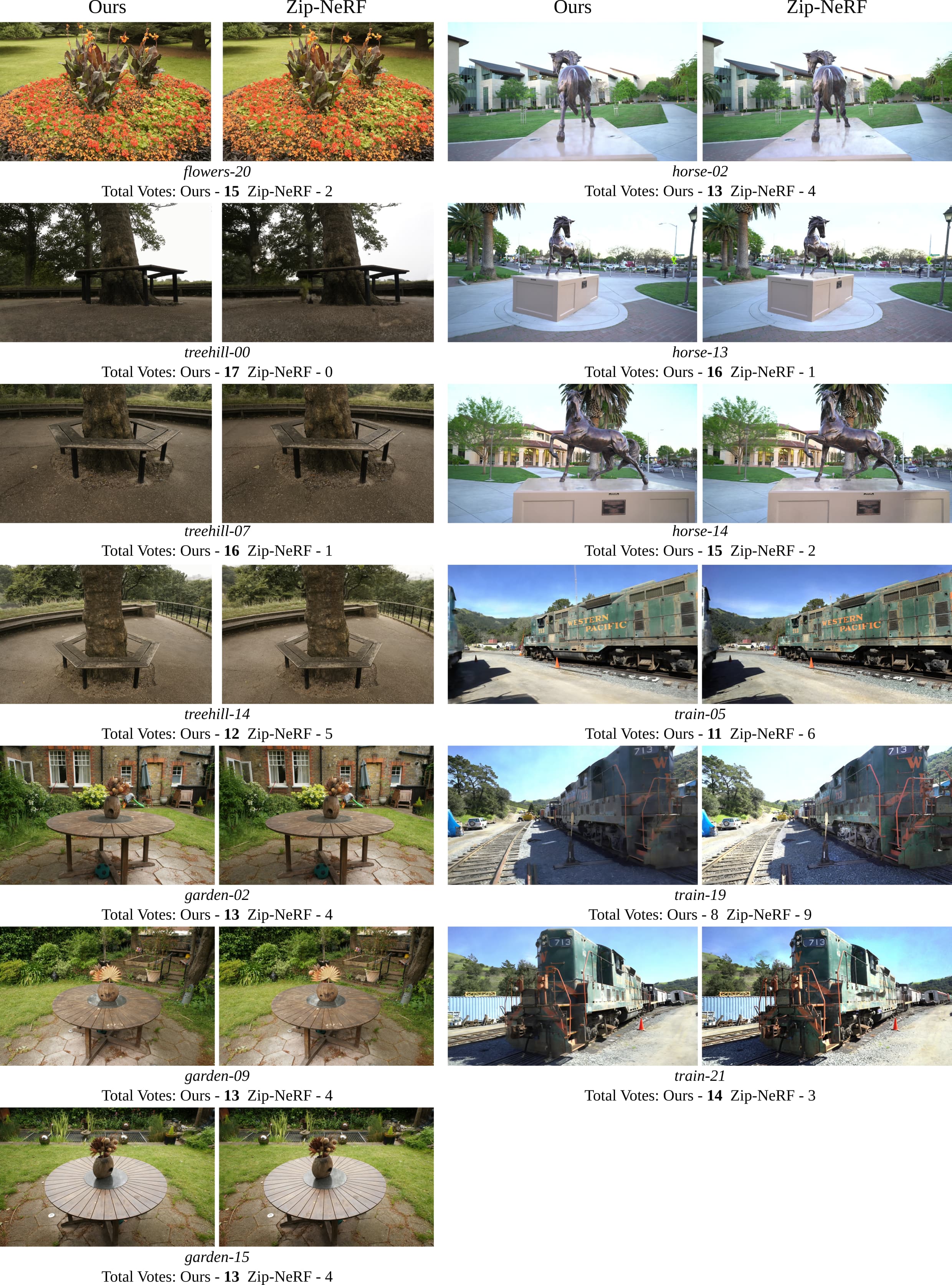}
\label{fig:exp3}%
\end{figure*}

\end{document}